\documentclass[10pt,journal,compsoc]{IEEEtran}

\usepackage{times}
\usepackage{epsfig}
\usepackage{graphicx}
\usepackage{amsmath}
\usepackage{amssymb}
\usepackage{url}
\usepackage{amsfonts}
\usepackage{xcolor}
\usepackage{bm}
\usepackage{multirow}
\usepackage{colortbl}
\usepackage{hhline}
\usepackage{pifont}
\usepackage{mathtools}
\usepackage{wrapfig}
\usepackage{cite}
\usepackage[caption=false]{subfig}
\usepackage{ragged2e}
\usepackage{makecell}
\usepackage[pagebackref=true,breaklinks=true,colorlinks,bookmarks=false]{hyperref}

\graphicspath{{./Imgs/}}
\DeclareGraphicsExtensions{.pdf,.jpg,.png}

\newcommand{\myPara}[1]{\vspace{.01in}\noindent\textbf{#1}\quad}
\definecolor{mygray}{gray}{.92}

\newcommand{\figref}[1]{Fig.~\ref{#1}}
\newcommand{\tabref}[1]{Tab.~\ref{#1}}
\newcommand{\secref}[1]{\S\ref{#1}}

\newcommand{\equref}[1]{Eq.~(\ref{#1})}

\def\ie{\emph{i.e.}}
\def\eg{\emph{e.g.}}

\def\etal{{\em et al.~}}

\DeclarePairedDelimiter\floor{\lfloor}{\rfloor}

\begin{document}

%%%%%%%%% TITLE
\title{Learning Local and Global Temporal Contexts for Video Semantic Segmentation}

\markboth{IEEE TRANSACTIONS ON PATTERN ANALYSIS AND MACHINE INTELLIGENCE}%
{Learning Local and Global Temporal Contexts for Video Semantic Segmentation}

\author{
  Guolei Sun, Yun Liu$^\dagger$, Henghui Ding, Min Wu, and Luc Van Gool
  \IEEEcompsocitemizethanks{%
    \IEEEcompsocthanksitem G. Sun and L. V. Gool are with Computer Vision Lab, ETH Zurich, Zurich, Switzerland.
    \IEEEcompsocthanksitem Y. Liu and M. Wu are with the Institute for Infocomm Research (I2R), Agency for Science, Technology and Research (A*STAR), Singapore. 
    \IEEEcompsocthanksitem H. Ding is with MMLab@NTU, Nanyang Technological University, Singapore.
    \IEEEcompsocthanksitem A preliminary version of this work has been published on CVPR 2022~\cite{sun2022coarse}.
    \IEEEcompsocthanksitem Corresponding author: Yun Liu (E-mail: VAGRANTLYUN@GMAIL.COM)
  }
}

\IEEEtitleabstractindextext{%
%%%%%%%%% ABSTRACT
\begin{abstract} \justifying
Contextual information plays a core role for video semantic segmentation (VSS). This paper summarizes contexts for VSS in two-fold: \textit{local} temporal contexts (LTC) which define the contexts from neighboring frames, and \textit{global} temporal contexts (GTC) which represent the contexts from the whole video. As for LTC, it includes static and motional contexts, corresponding to static and moving content in neighboring frames, respectively. Previously, both static and motional contexts have been studied. However, there is no research about simultaneously learning static and motional contexts (highly complementary). Hence, we propose a Coarse-to-Fine Feature Mining (CFFM) technique to learn a unified presentation of LTC. CFFM contains two parts: Coarse-to-Fine Feature Assembling (CFFA) and Cross-frame Feature Mining (CFM). CFFA abstracts static and motional contexts, and CFM mines useful information from nearby frames to enhance target features. To further exploit more temporal contexts, we propose CFFM++ by additionally learning GTC from the whole video. Specifically, we uniformly sample certain frames from the video and extract global contextual prototypes by $k$-means. The information within those prototypes is mined by CFM to refine target features. Experimental results on popular benchmarks demonstrate that CFFM and CFFM++ perform favorably against state-of-the-art methods. The code is available at \href {https://github.com/GuoleiSun/VSS-CFFM}{https://github.com/GuoleiSun/VSS-CFFM}.
\end{abstract}
}

\maketitle

\IEEEdisplaynontitleabstractindextext
\IEEEpeerreviewmaketitle

%%%%%%%%% BODY TEXT
\section{Introduction}\label{sec:introduction}
\IEEEPARstart{S}{emantic} segmentation aims at assigning a semantic label to each pixel in a natural image, which is a fundamental and hot topic in the computer vision community. It has a wide range of applications in both academic and industrial fields. Thanks to the powerful representation capability of deep neural networks \cite{shelhamer2017fully,badrinarayanan2017segnet,tian2019decoders,liu2020efficientfcn} and large-scale image datasets \cite{everingham2015pascal,cordts2016cityscapes,zhou2019semantic,caesar2018coco,waqas2019isaid}, tremendous achievements have been seen for image semantic segmentation. However, \textbf{video semantic segmentation (VSS)} has not been witnessed such tremendous progress \cite{gadde2017semantic,jin2017video,liu2017surveillance,nilsson2018semantic} due to the lack of large-scale datasets. For example, Cityscapes \cite{cordts2016cityscapes} and NYUDv2 \cite{silberman2012indoor} datasets only annotate one or several nonadjacent frames in a video clip. CamVid \cite{brostow2008segmentation} only has a small scale and a low frame rate. The real world is actually dynamic rather than static, so research on VSS is necessary. Fortunately, the recent establishment of the large-scale VSS dataset, VSPW \cite{miao2021vspw}, solves the problem of video data scarcity. This inspires us to denote our efforts to VSS.

\begin{figure}
\centering
\subfloat[]{\label{fig:intro_a}\includegraphics[width=0.9\linewidth]{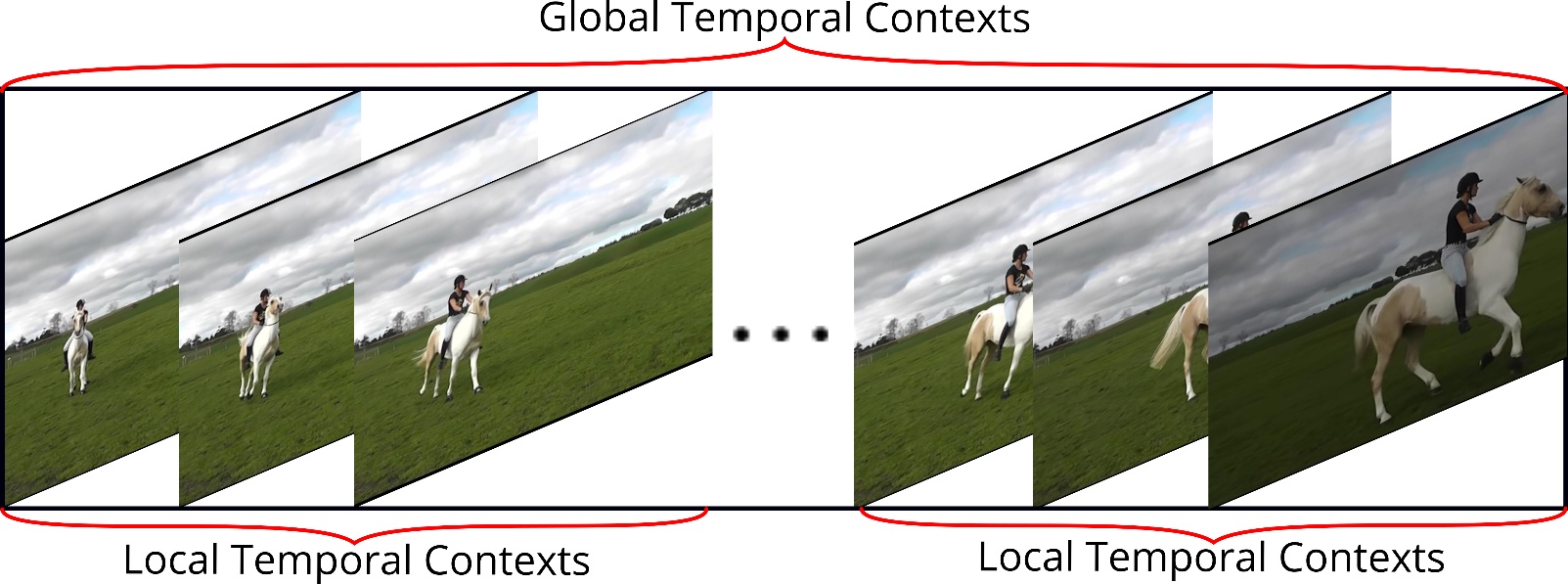}} \\
\subfloat[]{\label{fig:intro_b}\includegraphics[width=0.9\linewidth]{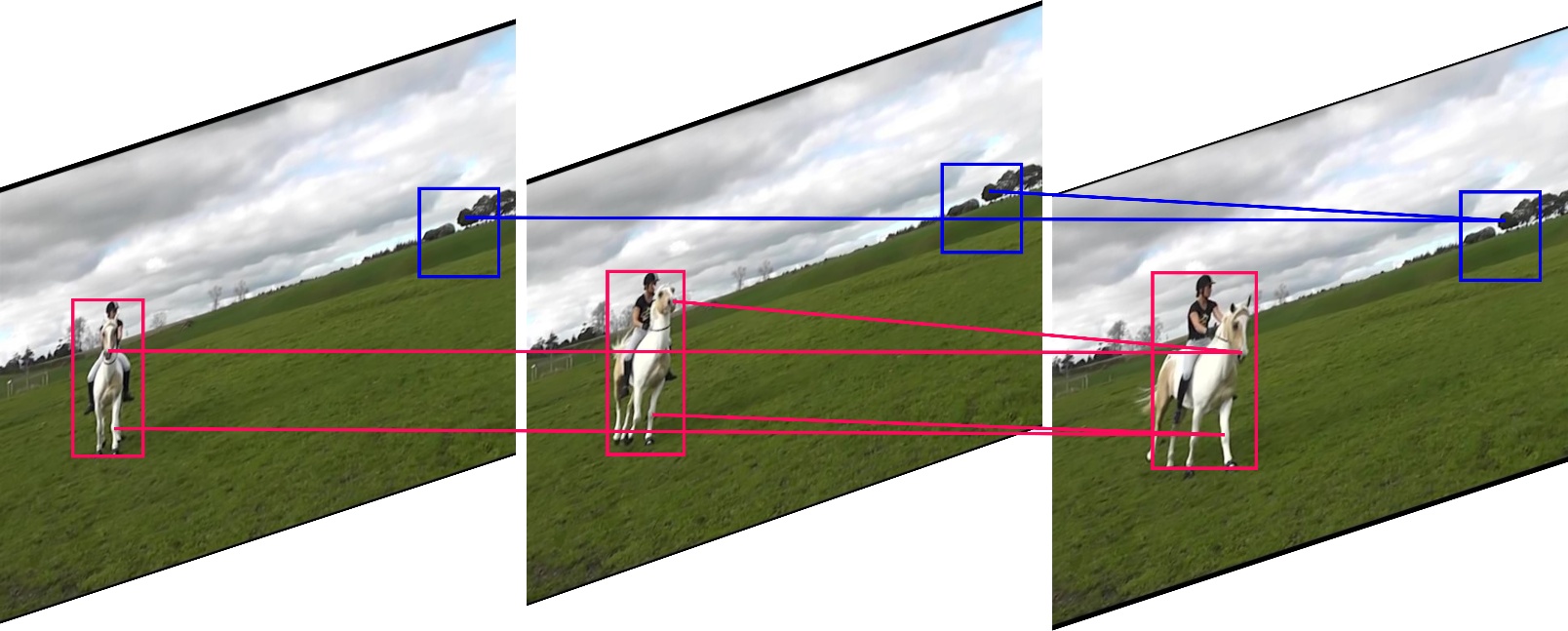}}
\caption{\textbf{Illustration of various video contexts.} (a) Illustration of \textit{local temporal contexts} and \textit{global temporal contexts}. (b) Illustration of \textit{static contexts} (in blue) and \textit{motional contexts} (in red) across neighbouring video frames. The human and horse are moving objects, while the grassland and sky are static backgrounds. Note that the static stuff is helpful for the recognition of moving objects, \ie, a human is riding a horse on the grassland.}
\label{fig:intro}
\vspace{-3mm}
\end{figure}

As widely accepted, the contextual information plays a central role in image semantic segmentation \cite{zhang2018context,jin2021mining,jin2021isnet,yuan2020object,liu2020learning,zhou2019context,he2019adaptive,ding2018context,li2020spatial,chen2018deeplab,chen2018encoder,zhao2017pyramid,yang2018denseaspp,huang2019ccnet,zhu2019asymmetric,zhen2020joint}. When considering videos, the contextual information can be divided into two cases based on how much temporal information is used: \textbf{\textit{local temporal contexts}} and \textbf{\textit{global temporal contexts}}. As shown in \figref{fig:intro_a}, local temporal contexts refer to the contexts from neighboring/nearby frames, while global temporal contexts represent the contexts from a much larger view, \ie, the whole video.

We first discuss local temporal contexts which are widely exploited in VSS~\cite{gadde2017semantic,nilsson2018semantic,liu2017surveillance,jin2017video,xu2018dynamic,liu2020efficient,hu2020temporally,shelhamer2016clockwork,jain2019accel,carreira2018massively,zhu2017deep,li2018low,lee2021gsvnet}. The local temporal contexts can be further divided into \textbf{\textit{static contexts}} and \textbf{\textit{motional contexts}} among neighboring video frames, as shown in \figref{fig:intro_b}. The former refers to the contexts within the same video frame or the contexts of unchanged content across the neighboring frames. Image semantic segmentation has exploited such contexts (for images) a lot, mainly accounting for multi-scale \cite{chen2018deeplab,chen2018encoder,yang2018denseaspp,he2019adaptive} and global/long-range information \cite{zhao2017pyramid,zhang2018context,huang2019ccnet,zhu2019asymmetric}. Such information is essential not only for understanding the static scene but also for perceiving the relatively holistic environment existing in the neighboring frames. The latter is responsible for better parsing moving objects/stuff and capturing more effective scene representations with the help of motions. Most of the VSS methods mainly studied motional contexts among nearby frames, which usually relies on optical flows \cite{dosovitskiy2015flownet} to model motional contexts from frames to adjacent frames, ignoring the static contexts. Although each single aspect, \ie, static or motional contexts, has been well studied, how to learn static and motional contexts simultaneously among nearby frames deserves more attention, which is important for VSS.

Furthermore, static contexts and motional contexts are highly correlated, not isolated, because both contexts are complementary to each other to represent the information existing in several nearby frames. Therefore, the ideal solution for learning local temporal contexts is to jointly learn static and motional contexts, \ie, generating a unified representation of static and motional contexts. A na\"ive solution is to apply recent popular self-attention \cite{vaswani2017attention,dosovitskiy2021image,wang2018non} by taking feature vectors at all pixels in neighboring frames as tokens. This can directly model global relationships of all tokens, of course including both static and motional contexts. However, this na\"ive solution has some obvious drawbacks. For example, it is super inefficient due to a large number of tokens/pixels in the considered nearby frames, making this na\"ive solution unrealistic. It also contains too much redundant computation because most content in nearby frames usually does not change much and it is unnecessary to compute attention for the repeated content. Moreover, the too-long length of tokens would affect the performance of self-attention, as shown in \cite{xu2021co,liu2021swin,heo2021rethinking,fan2021multiscale,chu2021twins} where the reduction of the token length through downsampling leads to better performance. More discussion about why traditional self-attention is inappropriate for video context learning can be found in \secref{sec:ltc_motivation}.

In this paper, we propose a \textbf{Coarse-to-Fine Feature Mining (CFFM)} technique to learn local temporal contexts, which consists of two parts: \textbf{Coarse-to-Fine Feature Assembling (CFFA)} and \textbf{Cross-frame Feature Mining (CFM)}. Specifically, we first apply an efficient deep network \cite{xie2021segformer} to extract features from each frame. Then, we assemble the extracted features from neighboring frames in a coarse-to-fine manner. Here, we use a larger receptive field and a more coarse pooling if the frame is more distant from the target frame. This feature assembling operation has two meanings. On one hand, it organizes the features in a multi-scale way, and the farthest frame would have the largest receptive field and the most coarse pooling. Since the content in a few sequential frames usually does not change suddenly and most content may only have a little temporal inconsistency, this operation is expected to prepare data for learning static contexts. On the other hand, this feature assembling operation enables a large perception region for remote frames because the moving objects may appear in a large region for remote frames. This makes it suitable for learning motional contexts. Then, with the assembled features, we use the CFM technique to iteratively mine useful contextual information from neighbouring frames for the target frame. This mining technique is a specially designed non-self attention mechanism that has two different inputs, unlike commonly used self-attention that only has one input \cite{vaswani2017attention,dosovitskiy2021image}. The output features enhanced by CFFM can be directly used for final prediction. We describe the technical motivations for CFFM in detail in \secref{sec:ltc_motivation}.

For global temporal contexts, few VSS methods~\cite{miao2021vspw,zhang2022auxadapt} have exploited the contexts from the whole video. The modeling of global temporal contexts is usually achieved by a memory module in the form of a memory bank~\cite{miao2021vspw} or a tiny network~\cite{zhang2022auxadapt} which is updated during inference. Although promising results have been achieved, there are two obvious drawbacks: 1) the global temporal contexts are \textit{implicitly} modeled and it is unclear what information is kept in the memory; 2) the contextual information in the memory keeps increasing when processing the video frame-by-frame and the global temporal interaction is only possible for the last few frames of the video. To this end, based on the proposed CFFM, we further propose to \textit{explicitly} learn global temporal contexts for VSS. After training our CFFM, the features (for each frame of the video) output from the trained network contains high-level semantic information and can be used to extract global temporal contexts. Since a video contains a large number of frames (tens or hundreds), we first sample some frames by a certain step from the whole video. This largely reduces the number of frames for the following processing. Features are extracted for these selected frames, which are decomposed as tokens. Here, the number of tokens is still large and impossible to be used. We largely reduce the token quantity by clustering all the tokens into different sub-groups. The centers of sub-groups are informative and representative prototypes, which abstract the contexts for the whole video. With the generated prototypes, we use the CFM technique again to iteratively mine useful information from the whole video to the target frame. The prediction of this global temporal context mining branch is combined with the prediction from CFFM. We name this model using both local and global temporal contexts as CFFM++, which is an extension of the CFFM by incorporating global temporal contexts.

To summarize, this paper studies local and global temporal contexts for VSS, with the following contributions:
\begin{itemize}
\itemsep0em 
\item To learn the \textit{local temporal contexts} of videos, we propose CFFM technique to learn a unified representation of \textit{static contexts} and \textit{motional contexts} among neighboring video frames, both of which are of vital importance for VSS.
\item To learn the \textit{global temporal contexts} of videos, we propose a global temporal context mining module to explicitly incorporate contextual information from the whole video to the target frame.
\item Without bells and whistles, we achieve state-of-the-art results for VSS on standard benchmarks by using the CFFM technique. With the global temporal contexts incorporated, CFFM++ further boosts the performance of VSS.
\end{itemize}

We build this paper upon our recent conference paper~\cite{sun2022coarse} and significantly extend it in various ways. First, we propose an extension method CFFM++ (\figref{fig:framework_exten}) based on the original framework (CFFM) to exploit \textit{global temporal contexts} from the whole video, further boosting the segmentation performance while introducing only limited computation. Second, we provide more in-depth discussions on motivations, related works, and implementation (\secref{sec:introduction}, \secref{sec:gtc}, and \secref{sec:experi}). Third, we conduct more ablation studies to thoroughly examine each key component of the proposed method, on top of which we provide more insights (\secref{sec:experi}). Fourth, extensive experiments on two challenging datasets are performed to demonstrate the effectiveness of learning global temporal contexts (\secref{sec:experi}). Last but not least, we provide more visual results (\figref{fig:qual}) to better show the advantages of CFFM and CFFM++.

\section{Related Work}
\subsection{Image Semantic Segmentation}
Image semantic segmentation has always been a key topic in the vision community, mainly because of its wide applications in real-world scenarios. Since the pioneer work of FCN \cite{shelhamer2017fully} which adopts fully convolution networks to make densely pixel-wise predictions, a number of segmentation methods have been proposed with different motivations or techniques \cite{zhang2021self,hsiao2021specialize,zhu2021learning,liu2021exploit,wang2020dual,pang2019towards,nirkin2021hyperseg,hu2020class,chen2020tensor,zhang2019co,wei2019building}. For example, some works try to design effective encoder-decoder network architectures to exploit multi-level features from different network layers \cite{badrinarayanan2017segnet,shelhamer2017fully,chen2018encoder,tian2019decoders,liu2020efficientfcn}. Some works impose extra boundary supervision to improve the prediction accuracy of details \cite{yu2018learning,takikawa2019gated,zhen2020joint,liu2022semantic,li2020improving,wang2021active}. Some works utilize the attention mechanism to enhance the semantic representations \cite{chen2016attention,li2019expectation,zhang2019acfnet,fu2019dual,seifi2020attend,zhong2020squeeze,huang2019ccnet,zhu2019asymmetric}. Besides these talent works, we want to emphasize that most research aims at learning powerful contextual information \cite{zhang2018context,yuan2020object,liu2020learning,zhou2019context,ding2019semantic,he2019adaptive,zhen2020joint,li2020spatial,jin2021isnet,jin2021mining}, including multi-scale \cite{chen2018deeplab,chen2018encoder,yang2018denseaspp,he2019adaptive,he2019dynamic,li2020spatial} and global/long-range information \cite{zhao2017pyramid,zhang2018context,huang2019ccnet,zhu2019asymmetric}. The contextual information is also essential for VSS, but video contexts are different from image contexts, as discussed above.

\subsection{Video Semantic Segmentation}
Since the real world is dynamic rather than static, VSS is necessary for pushing semantic segmentation into more practical deployments. Previous research on VSS was limited by the available datasets \cite{miao2021vspw}. Specifically, three datasets were available: Cityscapes \cite{cordts2016cityscapes}, NYUDv2 \cite{silberman2012indoor}, and CamVid \cite{brostow2008segmentation}.
% Cityscapes \cite{cordts2016cityscapes} and NYUDv2 \cite{silberman2012indoor} only annotate several nonadjacent frames in a video clip. CamVid \cite{brostow2008segmentation} only has a small scale, a low frame rate (1fps), and low resolution (360 $\times$ 480). 
They either only annotate several nonadjacent frames in a video clip or have a small scale, a low frame rate, and low resolution.
Fortunately, the recent establishment of the large-scale, fully-annotated VSPW dataset \cite{miao2021vspw} solves this problem.

Most of the existing VSS methods utilize the optical flow to capture temporal relations \cite{gadde2017semantic,nilsson2018semantic,liu2017surveillance,liu2020efficient,shelhamer2016clockwork,jain2019accel,zhu2017deep,xu2018dynamic,lee2021gsvnet,kundu2016feature,mahasseni2017budget}. These methods usually adopt different smart strategies to balance the trade-off between accuracy and efficiency \cite{kundu2016feature,mahasseni2017budget}. Among them, some works aim at improving the segmentation accuracy by exploiting the temporal relations using the optical flow for feature warping \cite{gadde2017semantic,liu2017surveillance,nilsson2018semantic} or the GAN-like architecture \cite{goodfellow2014generative} for predictive feature learning \cite{jin2017video}. The other works aim at improving the segmentation efficiency by using temporal consistency for feature propagation and reuse \cite{jain2019accel,li2018low,shelhamer2016clockwork,zhu2017deep}, or directly reusing high-level features \cite{shelhamer2016clockwork,carreira2018massively}, or adaptively selecting the key frame \cite{xu2018dynamic}, or propagating segmentation results to neighbouring frames \cite{lee2021gsvnet}, or extracting features from different frames with different sub-networks \cite{hu2020temporally}, or considering the temporal consistency as extra training constraints \cite{liu2020efficient}. Zhu \etal \cite{zhu2019improving} utilized video prediction models to predict future frames as well as future segmentation labels, which are used as augmented data for training better image semantic segmentation models, not for VSS. Different from the above approaches, STT~\cite{li2021video} and LMANet~\cite{paul2021local} directly model the interactions between the target and reference frame features to exploit the temporal information.

The above VSS approaches explore the local temporal relation, here denoted as \textit{motional contexts}. However, \textit{local temporal contexts} include two aspects: \textit{static and motional contexts}. Those methods ignore the static contexts that are important for segmenting complicated scenes. This paper addresses this problem by proposing a new video context learning mechanism, capable of learning a unified representation of static and motional contexts. Besides, we also propose to explicitly learn \textit{global temporal contexts} with prototype learning and attention-based feature mining.

\subsection{Difference with STT}
We notice that a concurrent work STT~\cite{li2021video} also utilizes bigger searching regions for more distant video frames and self-attention for establishing connections across frames. While the two works share these similarities, there are key differences between them. {\textit{First}}, two methods have different motivations. We target exploiting both static and motional contexts (local temporal contexts), while STT focuses on capturing the temporal relations among complex regions. Note that the concept of static/motional contexts is similar to the concept of simple/complex regions in STT. As a result, STT models only the motional contexts, while our method models both static and motional contexts. {\textit{Second}}, the designs are different. For query selection, STT selects 50\% of query locations in order to reduce the computation. However, our method splits the query features into windows and the query features in each window share the same contexts to reduce the computation. For key/value selection, STT operates in the same granularity, while our method processes the selected key/value into different granularity, which reduces the number of tokens and models the multi-scale information for static contexts. {\textit{Third}}, our cross-frame feature mining can exploit multiple transformer layers to deeply mine the contextual information from the reference frames, but STT only uses one layer. The reason may be that STT only updates the query features of the selected locations and using multiple STT layers could lead to inconsistency in the query features in un-selected and selected locations. Moreover, this paper also exploits global temporal contexts for further improvement.

\subsection{Vision Transformer}
Vision transformer, a strong competitor of convolutional neural networks (CNNs), has been widely adopted in various vision tasks~\cite{dosovitskiy2021image,liu2021swin,MeViS,yang2021focal,liang2021swinir,yuan2021tokens,liu2021vision,SETR,MOSE,yu2021pointr,sun2023rethinking}, due to its powerful ability of modeling global connection within all the input tokens. Specifically, ViT~\cite{dosovitskiy2021image} splits an image into patches to construct tokens and processes tokens using typical transformer layers. Swin Transformer~\cite{liu2021swin} improves ViT by introducing shifted windows when computing self-attention. 
% Focal Transformer~\cite{yang2021focal} introduces both fine-grained and coarse-grained attention in architecture design.
The effectiveness of transformers has been validated in tracking~\cite{chen2021transformer,yan2021learning}, crowd counting~\cite{liang2021transcrowd,sun2023rethinking}, multi-label classification~\cite{lanchantin2021general} and so on. In the following, we specifically discuss the transformer-based segmentation methods.

To improve segmentation using transformers, some methods~\cite{SETR,wang2021max,xie2021segformer,VLTTPAMI,li2022panoptic,cheng2021per,cheng2022masked} have been developed. SETR~\cite{SETR} and Panoptic SegFormer~\cite{li2022panoptic} are the first transformer-based models for image and panoptic semantic segmentation, respectively. Generally, these works use transformers to generate global-context-aware features. Differently, a new trend of works such as MaskFormer~\cite{cheng2021per} and Mask2Former~\cite{cheng2022masked} use transformer decoders to get rid of the conventional per-pixel classification for segmentation. ViT-Adapter~\cite{chen2023vision} learns powerful representations from large-scale multi-modal data and allows plain ViT to achieve comparable performance to vision-specific transformers. For video understanding, \cite{zhang20231st} and \cite{su20233rd} exploit transformers to merge temporal information and achieve promising results on the video panoptic segmentation task.
Despite the success of transformers in segmentation, the use of transformer layers in VSS is non-trivial due to the large number of tokens from video frames. Here, we propose an effective and efficient way to model the temporal contextual information for VSS. A concurrent work MRCFA~\cite{sun2022mining} also works on VSS using transformers. However, MRCFA specifically focuses on refining feature affinity maps, while this paper focuses on learning local and global temporal contexts for the video.

In terms of designing vision transformers to use contextual information, Focal Transformer~\cite{yang2021focal} introduces both fine-grained and coarse-grained attention in architecture design to explore local and global contexts in the image. Though our proposed methods also focus on learning contexts, there are significant differences. First, our methods focus on \textit{video contexts}, while the Focal Transformer explores \textit{image contexts}. Video contexts are much more complex than the image contexts. As illustrated in Fig.~\ref{fig:intro}, video contexts include \textit{local temporal} contexts which represent the contexts from neighboring frames, and \textit{global temporal} contexts which mean the contexts from the whole video. For local temporal contexts, they can be further divided into \textit{static} and \textit{motional} contexts. However, the image contexts studied in Focal Transformer only refer to the information from a local region of the image or the global region (the whole image). Second, adding a new temporal dimension makes the problem of learning contexts much more challenging and significantly increases implementation difficulty. Third, our methods specifically focus on the VSS task, which are built upon a pre-trained backbone/encoder. We achieve promising performance on popular VSS datasets. Differently, the Focal Transformer proposes a new network architecture/encoder and focuses on image understanding tasks such as image classification, detection, and segmentation.

\begin{figure*}[!t]
    \centering
    \includegraphics[width=\linewidth]{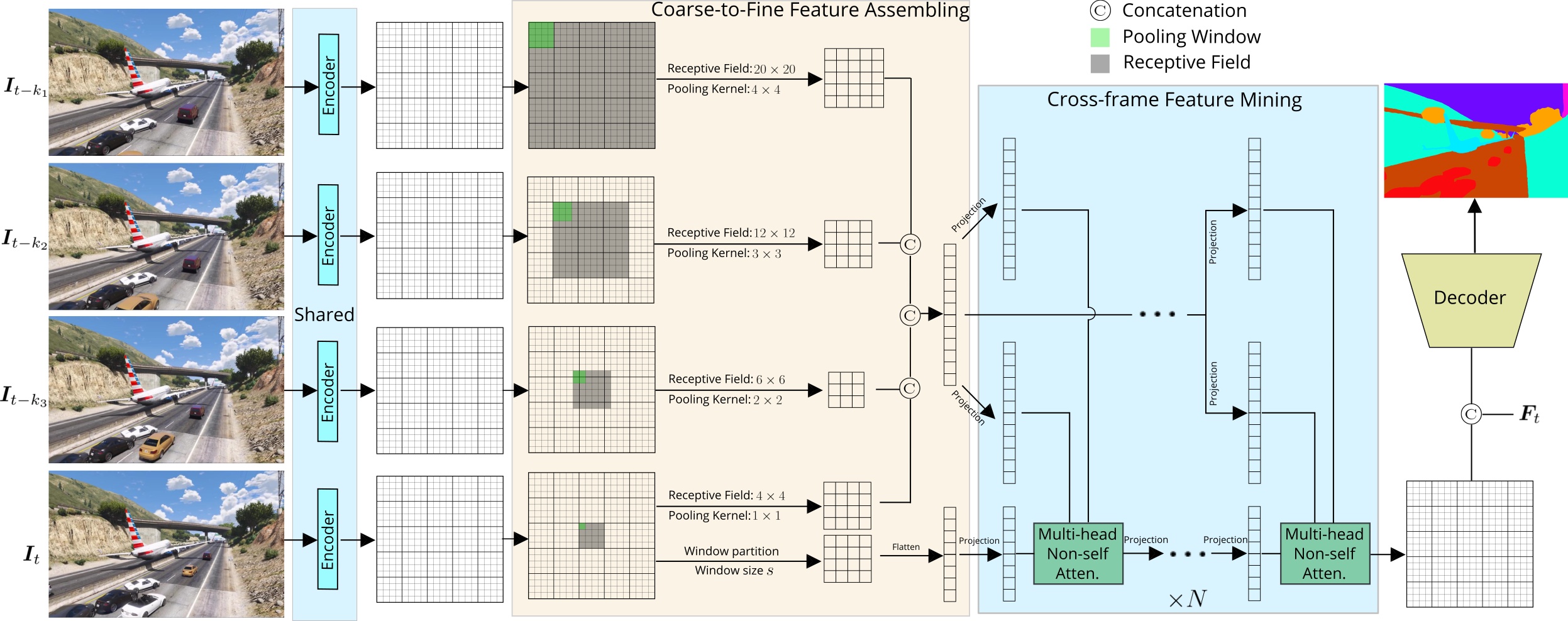}
    \caption{{\textbf{Overview of the proposed Coarse-to-Fine Feature Mining for mining local temporal contexts.}} All frames are first input to an encoder to extract features, which then go through the coarse-to-fine feature assembling module (CFFA). Features for different frames are processed by different pooling strategies to generate the context tokens. The principle is that for more distant frames, a bigger receptive field and more coarse pooling are used. The shown feature size ($20 \times 20$), receptive field, and pooling kernel are for a simple explanation. The context tokens from all frames are concatenated and then processed by the cross-frame feature mining (CFM) module. The context tokens are exploited to update the target features by several multi-head non-self attention layers. Finally, we use the enhanced target features to make the segmentation prediction for the target frame. \textit{Best viewed with zooming.}}
    \label{fig:framework}
\end{figure*}

\section{Local Temporal Contexts}\label{sec:ltc}
In this section, we focus our discussion on local temporal contexts. To begin with, we introduce the technical motivation of the proposed Coarse-to-Fine Feature Mining (CFFM) for mining the local temporal contexts in \secref{sec:ltc_motivation}. Then, we introduce the first sub-operation of Coarse-to-Fine Feature Assembling (CFFA) in \secref{sec:assembling}. Next, we present the second sub-operation of Cross-frame Feature Mining (CFM) in \secref{sec:mining}. At last, we analyze the complexity in \secref{sec:complexity}.

\subsection{Technical Motivation}\label{sec:ltc_motivation}
Before introducing our method, we discuss our technical motivation to help readers better understand the proposed technique. As discussed above, local temporal contexts include static contexts and motional contexts. The former has been well exploited in image semantic segmentation \cite{zhang2018context,jin2021mining,jin2021isnet,yuan2020object,liu2020learning,zhou2019context,ding2019boundary,he2019adaptive,he2019dynamic,li2020spatial,chen2018deeplab,chen2018encoder,zhao2017pyramid,yang2018denseaspp,huang2019ccnet,zhu2019asymmetric,zhen2020joint}, while the latter has been studied in VSS \cite{li2018low,carreira2018massively,gadde2017semantic,nilsson2018semantic,liu2017surveillance,liu2020efficient,shelhamer2016clockwork,jain2019accel,zhu2017deep,xu2018dynamic,lee2021gsvnet,kundu2016feature,mahasseni2017budget}. However, there is no research touching the joint learning of both static and motional contexts which are both essential for VSS.

To address this problem, a na\"ive solution is to simply apply the recently popular self-attention mechanism \cite{vaswani2017attention,dosovitskiy2021image,wang2018non} to the video sequence by viewing the feature vector at each pixel of each frame as a token. In this way, we can model global relationships by connecting each pixel with all others, so all local temporal contexts can of course be constructed. However, this na\"ive solution has \textit{three obvious drawbacks}. First, a video sequence has $l+1$ times more tokens than a single image, where $l+1$ is the length of the video sequence. This would lead to $(l+1)^2$ times more computational cost than a single image because the complexity of the self-attention mechanism is $\mathcal{O}(n^2c)$, where $n$ is the number of tokens and $c$ is the feature dimension \cite{vaswani2017attention,dosovitskiy2021image,liu2021swin}. Such high complexity is unaffordable, especially for VSS which needs on-time processing as the video data stream comes in sequence. Second, such direct global modeling would be redundant. Despite that there are some motions in a video clip, the overall semantics/environment would not change suddenly and most video content is repeated. Hence, most of the (self-to-self) connections built by direct global modeling are unnecessary. Last but not least, although self-attention can technically model global relationships, a too-long sequence length would limit its performance, as demonstrated in \cite{wang2021pyramid,xu2021co,liu2021swin,heo2021rethinking,fan2021multiscale,chu2021twins} where downsampling features into small scales leads to better performance than the original long sequence length.

Instead of directly modeling global relationships, we propose to model relationships only among necessary tokens for the joint learning of static and motional contexts. Our CFFM technique consists of two steps. The first step, Coarse-to-Fine Feature Assembling (CFFA), assembles the features extracted from neighbouring frames in a temporally coarse-to-fine manner based on \textit{three observations}. First, the moving objects/stuff can only move gradually across frames in practice, and the objects/stuff cannot move from one position to another far position suddenly. Thus, the region of the possible positions of (a) moving object/stuff in a frame gradually gets larger for farther frames. In other words, for one pixel in a frame, the farther the frames, the larger the correlated regions. Second, although some content may change across frames, the overall semantics and environment would not change much, which means that most video content may only have a little temporal inconsistency. For statistical evidence, we compute the mIoU between the ground-truth masks of consecutive video frames on the VSPW val set~\cite{miao2021vspw}, to show that the semantic masks for consecutive frames are largely overlapped and the scene changes are thus very small from a frame to its next frame. The obtained mIoU is 89.7\%, proving that the objects/background move slowly from frame to frame. Third, the little temporal inconsistency of the ``static'' content across neighbouring frames can be easily handled by the pooling operation which is scale- and rotation-invariant, as evidenced in previous works~\cite{shelhamer2017fully,zhao2017pyramid,zhang2018context,liu2020learning}. Inspired by the second and third observations, a varied-size region sampling through the pooling operation in neighbouring frames can convey multi-scale contextual information. Therefore, the designed CFFA can perceive multi-scale contextual information (static contexts) and motional contexts. Specifically, each pixel in the target frame corresponds to a larger receptive field and a more coarse pooling in the farther frame, as depicted in \figref{fig:framework}. Note that the length of the sampled tokens is much shorter than that in the default self-attention.

The second step of CFFM, Cross-frame Feature Mining (CFM), is designed to mine useful information from the features of neighbouring frames. This is an attention-based process. However, unlike traditional self-attention \cite{vaswani2017attention,dosovitskiy2021image,wang2018non} whose query, key, and value come from the same input, we propose to use a \textit{non-self attention} mechanism, where the query is from the target frame and the key and value are from neighbouring frames. Besides, we only update the query during the iterative running of non-self attention, but we keep the context tokens unchanged. This is intuitive as our goal is to mine information from neighbouring frames and the update of context tokens is thus unnecessary. Compared with self-attention which needs to process all assembled features, this non-self attention further reduces the computational cost.

\subsection{Coarse-to-Fine Feature Assembling}
\label{sec:assembling}
Without loss of generalizability, we start our discussion on training data containing nearby video frames $\{\bm{I}_{t-k_1}, \cdots, \bm{I}_{t-k_l}, \bm{I}_{t} \}$ with ground-truth masks of $\{\bm{S}_{t-k_1}, \cdots, \bm{S}_{t-k_l}, \bm{S}_{t} \}$, and we focus on segmenting $\bm{I}_{t}$. Specifically, $\bm{I}_{t}$ is the target frame and $\{\bm{I}_{t-k_1}, \cdots, \bm{I}_{t-k_l}\}$ are $l$ previous reference frames which are $\{{k_1}, \cdots, {k_l}\}$ frames away from $\bm{I}_{t}$, respectively. Here, the local temporal contexts are considered since the reference frames are close to the target one. Let us denote $U = \{t-k_1, \cdots, t-k_l, t\}$ as the set of frame subscripts.
We first process $\{\bm{I}_{t-k_1}, \cdots, \bm{I}_{t-k_l}, \bm{I}_{t} \}$ using an encoder to extract informative features $\{\bm{F}_{t-k_1}, \cdots, \bm{F}_{t-k_l}, \bm{F}_{t} \}$, each of which has the size of $\mathbb{R}^{h \times w \times c}$ ($h$, $w$, and $c$ represent height, width, and feature dimension, respectively). We aim to exploit the features from the nearby frames to generate better features for segmenting $\bm{I}_{t}$ as valuable local temporal contexts exist in previous frames.

To efficiently establish long-range interactions between the reference frame features ($\{\bm{F}_{t-k_1}, \cdots, \bm{F}_{t-k_l}\}$) and the target frame features $\bm{F}_{t}$, we propose the coarse-to-fine feature assembling module, as shown in \figref{fig:framework}. Inspired by previous works~\cite{liu2021swin,wang2021pyramid,yang2021focal}, we split the target frame features $\bm{F}_{t}$ into windows and each window attends to a shared context token. The reason behind this is that attending each location in $\bm{F}_{t}$ to a specific context token requires huge computation and memory costs. When using window size of $s \times s$, $\bm{F}_{t}$ is partitioned into $\frac{h}{s}\times \frac{w}{s}$ windows. We obtain the new feature map $\bm{F}_{t}'$ as follows:
\begin{equation}
% \footnotesize
\begin{aligned}
    \bm{F}_{t} \in \mathbb{R}^{h \times w \times c} &\to \bm{F}'_{t} \in \mathbb{R}^{(\frac{h}{s} \times s) \times (\frac{w}{s} \times s) \times c} \\
    &\to \bm{F}'_{t} \in \mathbb{R}^{\frac{h}{s} \times \frac{w}{s} \times s \times s \times c}.
\end{aligned}
\end{equation}

Then, we generate context tokens from different frames. The main idea is to see a bigger receptive field and use a more coarse pooling if the frame is more distant from the target, which is why we call this step coarse-to-fine feature assembling. The motivation behind this is described in \secref{sec:ltc_motivation}. Formally, we define two sets of parameters: the receptive fields $r=\{r_{t-k_1}, \cdots, r_{t-k_l}, r_{t}\}$ and the pooling kernel/window sizes $p=\{p_{t-k_1}, \cdots, p_{t-k_l}, p_{t}\}$, when generating corresponding context tokens. For $t-k_1 < t-k_2 < \cdots < t-k_l < t$, we have $r_{t-k_1} \ge r_{t-k_2} \ge \cdots \ge r_{t-k_l} \ge r_{t}$ and $p_{t-k_1} \ge p_{t-k_2} \ge \cdots \ge p_{t-k_l} \ge p_{t}$. With this definition, we partition $\{\bm{F}_{t-k_1}, \cdots, \bm{F}_{t-k_l}, \bm{F}_t\}$ using pooling windows $p=\{p_{t-k_1}, \cdots, p_{t-k_l}, p_{t}\}$ to pool the features, respectively. The result is processed by a fully connected layer (FC) for dimension reduction. This is formulated as 
\begin{equation}\label{Eq:allpara_fc}
% \footnotesize
\begin{aligned}
    \bm{F}_{j} \in \mathbb{R}^{h \times w \times c} &\to \bm{E}_{j} \in \mathbb{R}^{\frac{h}{p_j} \times \frac{w}{p_j} \times (p_j \times p_j \times c)} \\
    &\stackrel{\rm FC}{\to} \bm{E}_{j} \in \mathbb{R}^{\frac{h}{p_j} \times \frac{w}{p_j} \times c},
\end{aligned}
\end{equation}
where $j\in U$. In \figref{fig:framework}, we have $r=\{20,12,6,4\}$ and $p=\{4,3,2,1\}$ for all frames (3 reference and 1 target frames). 
% Since we want to have the finest context on the target frame, $p_{t}$ is set to 1.

For each window partition $\bm{F}'_{t}[i] \in \mathbb{R}^{s\times s \times c}$ ($i \in \{1,2,\cdots,\frac{hw}{s^2}\}$) in the target features, we extract $\frac{r_{j}}{p_{j}}\times \frac{r_{j}}{p_{j}}$ elements from $\bm{E}_{j}$ around the area where the window lies in. This can be easily implemented using the \textit{unfold} function in PyTorch \cite{paszke2019pytorch}. Let $\bm{c}_{i,j}$ denote the obtained context tokens from $j$-th frame and for $i$-th window partition in the target features. We concatenate $\bm{c}_{i,j}$ into $\bm{c}_i$ as follows,
\begin{equation}
% \footnotesize
    \bm{c}_{i} = {\rm Concat}[\bm{c}_{i,j}],
\end{equation}
where $j\in U$, $\bm{c}_i \in \mathbb{R}^{m \times c}$ and $m=\sum_{j\in U}\frac{r^2_j}{p^2_j}$. The context tokens from the target frame are obtained by using the parameter set ($r_t$, $p_t$) to process the target features. In practice, we additionally use another parameter set ($r_t'$, $p_t'$) to generate more contexts from the target since the target features are more important. For simplicity, we focus our discussion by omitting ($r_t'$, $p_t'$) and using only ($r_t$, $p_t$) for the target.

To sum up, $\bm{c}_i$ contains the context information from all frames, which is used to refine the target frame features. As discussed in \secref{sec:ltc_motivation}, on one hand, $\bm{c}_i$ covers the tokens at possible positions where moving objects/stuff would appear, so it can be used for learning motional contexts. On the other hand, $\bm{c}_i$ is a multi-scale sampling of neighbouring frames with the temporal inconsistency solved by the pooling operation, so it can be used for learning static contexts.

\begin{figure*}[!t]
    \centering
    \includegraphics[width=\linewidth]{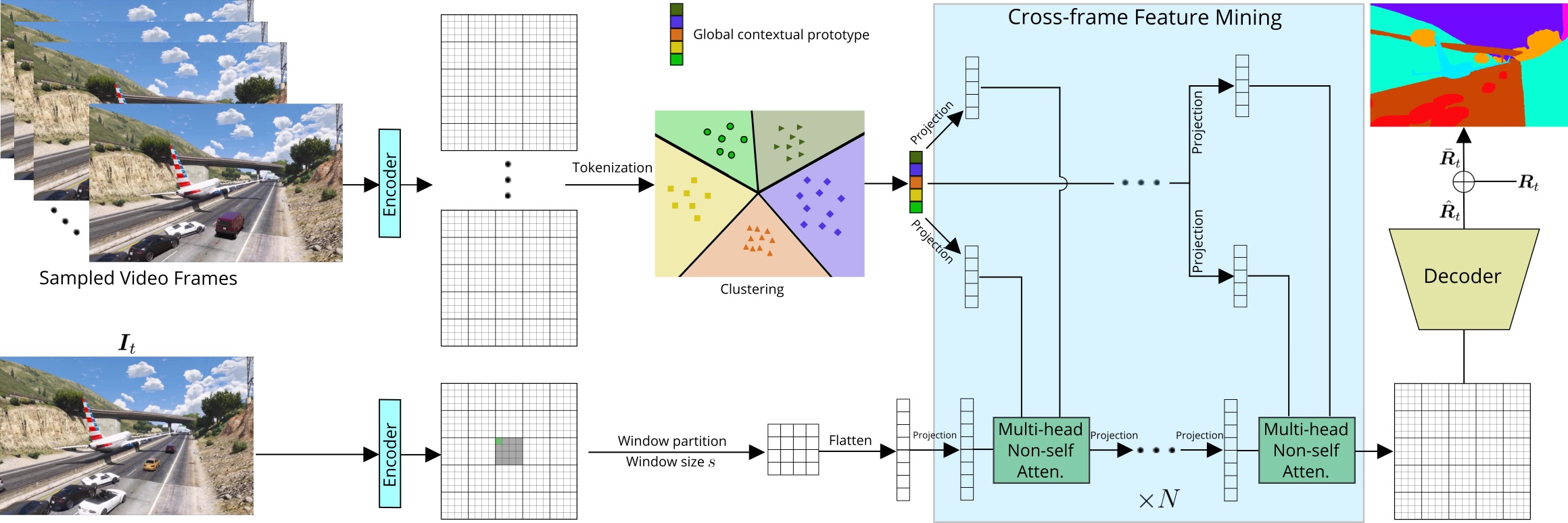}
    \caption{\textbf{Overview of the proposed CFFM++ for additionally mining global temporal contexts.} Due to the large number of frames in the video, we uniformly sample frames by a fixed step. The sampled video frames go through the encoder trained by CFFM and corresponding features are generated. After tokenizing the feature maps, we conduct unsupervised clustering ($k$-means) to reduce the tokens' number and learn global contextual prototypes. The obtained prototypes and the target frame features are passed to CFM, which enables the refinement of the target frame using {global temporal contexts}. The final predictions of CFFM++ are given by the weighted summation of the segmentation logits from learning local (CFFM) and global temporal contexts.}
    \label{fig:framework_exten}
\end{figure*}

\subsection{Cross-frame Feature Mining}\label{sec:mining}
After that we obtain the context token $\bm{c}_i$ for each window partition in the target features, we propose a non-self attention mechanism to mine useful information from neighboring frames. Unlike the traditional self-attention mechanism that computes the query, key, and value from the same input, our non-self attention mechanism utilizes different inputs to calculate the query, key, and value. Since $\bm{F}_{t}'$ is the input to the first layer of our CFM module, we re-write it as $\bm{F}^{0}_{t}=\bm{F}_{t}'$. For the $i$-th window partition in $\bm{F}^{0}_{t}$, the query $Q_i$, key $K_i$, and value $V_i$ are computed using three fully connected layers as follows:
\begin{align}\label{Eq:gene_qkv}
% \footnotesize
\begin{split}
    Q_i={\rm FC}(\bm{F}^{0}_{t}[i]),~~~K_i={\rm FC}(\bm{c}_i),~~~V_i={\rm FC}(\bm{c}_i),
\end{split}
\end{align}
where ${\rm FC}(\cdot)$ represents an FC layer. Next, we use non-self attention to update the target frame features, given by
\begin{align}\label{Eq:gene_attn}
% \footnotesize
\begin{split}
    \bm{F}^{1}_{t}[i]=\text{Softmax}(\frac{Q_i K_i^{T}}{\sqrt{c}}+B)V_i+\bm{F}^{0}_{t}[i],
\end{split}
\end{align}
where $B$ represents the position bias, following \cite{liu2021swin}. Note that we omit the formulation of the multi-head attention \cite{vaswani2017attention,dosovitskiy2021image} for simplicity. \equref{Eq:gene_qkv} and \equref{Eq:gene_attn} are repeated for $N$ steps, and we finally obtain the enhanced feature $\bm{F}^{N}_{t} \in \mathbb{R}^{\frac{h}{s} \times \frac{w}{s} \times s \times s \times c}$ for the target video frame. Local temporal contexts, \ie, static and motional contexts, from neighbouring video frames are continuously exploited to learn better representative features for segmenting the target frame. Note that in this process, we do not update the context tokens $\bm{c}_i$ for simplicity/elegance and reducing computation. Since this step is to mine useful information from the reference frames, it is also unnecessary to update $\bm{c}_i$. 
This is the advantage of non-self attention.

To generate segmentation predictions, we reshape $\bm{F}^{N}_{t}$ into $\mathbb{R}^{h \times w \times c}$ and concatenate $\bm{F}^{N}_{t}$ with $\bm{F}_{t}$. Then, a simple MLP projects the features to segmentation logits $\bm{R}_t$. The common cross-entropy loss (CE) is computed between $\bm{R}_t$ and ground-truth mask $\bm{S}_t$. Auxiliary losses on original features are also computed.
During inference, our method does not need to extract features for all $l+1$ frames when processing $\bm{I}_{t}$. Instead, the features of the reference frames, which are the frames before the target frame, have already been extracted in previous steps. Only the target frame is passed to the encoder to generate $\bm{F}_t$, and then features $\{\bm{F}_{t-k_1}, \cdots, \bm{F}_{t-k_l}, \bm{F}_t\}$ for all frames are passed to CFFM for representation enhancement.

\subsection{Complexity Analysis}\label{sec:complexity}
Here, we formally analyze the complexity of the proposed CFFM and the recent popular self-attention mechanism \cite{vaswani2017attention,dosovitskiy2021image,wang2018non} when processing video clip features $\{\bm{F}_{t-k_1}, \cdots, \bm{F}_{t-k_l}, \bm{F}_t\}$. The coarse-to-fine feature assembling (\equref{Eq:allpara_fc}) has the complexity of $\mathcal{O}((l+1)hwc)$, which is irrespective of $p$. The cross-frame feature mining has two parts: \equref{Eq:gene_qkv} has the complexity of $\mathcal{O}(hwc^2)+\mathcal{O}(mc^2)$, and \equref{Eq:gene_attn} is with the complexity of $\mathcal{O}(hwmc)$. As mentioned early, $m=\sum_{j\in U}\frac{r^2_j}{p^2_j}$. To sum over, the complexity of our method is given by
\begin{align}\label{Eq:gene_complexi}
% \scriptsize
\begin{split}
    \mathcal{O}({\rm CFFM})&=\mathcal{O}(hwmc)+\mathcal{O}(hwc^2)+\mathcal{O}(mc^2)\\
    &\ \ \ \ \ +\mathcal{O}((l+1)hwc)\\
    &=\mathcal{O}(hwmc)+\mathcal{O}(hwc^2),
\end{split}
\end{align}
where the derivation is conducted by removing less significant terms. For the self-attention mechanism \cite{vaswani2017attention,dosovitskiy2021image,wang2018non}, the complexity is $\mathcal{O}((l+1)^2h^2w^2c)+\mathcal{O}((l+1)hwc^2)$. Since $m\ll (l+1)^2hw$, the complexity of the proposed approach is much less than the self-attention mechanism. Take the example in Fig.~\ref{fig:framework}, $m=66$ while $(l+1)^2hw=6400$.

\section{Global Temporal Contexts}\label{sec:gtc}
In this section, we focus our discussion on the global temporal contexts. We start by explaining the process of extracting global temporal contextual information (prototypes). Then, we discuss how to exploit the generated contextual prototypes to refine the features of the target frame.

\subsection{Global Temporal Contextual Prototypes}
\label{sec:gtc_prototypes}
In the last section (\secref{sec:ltc}), we discuss how to learn local temporal contexts among nearby video frames $\{\bm{I}_{t-k_1}, \cdots, \bm{I}_{t-k_l}, \bm{I}_{t} \}$. Here, we propose to learn global temporal contexts to make the model have a much larger temporal view. To start with, we represent the corresponding whole video as $\bm{V}=\{\bm{I}_{1}, \cdots, \bm{I}_{t}, \cdots, \bm{I}_{T} \}$, containing a total of $T$ frames. Similar to \secref{sec:assembling}, we aim to segment the target frame $\bm{I}_t$ without losing generalizability. In the following, we explain the process of extracting the global temporal contextual information.

Our technique is built on the CFFM introduced in \secref{sec:ltc}. Once CFFM is trained, the encoder has the ability to generate informative features for each frame of the video. We first use the trained encoder to extract features for a subset of frames in the video $\bm{V}$. Specifically, the subset of frames are uniformly sampled from $\bm{V}$ by a fixed step $d$, which are denoted as $\bar{\bm{V}}=\{\bm{I}_{1}, \bm{I}_{1+d}, \cdots, \bm{I}_{1+(f-1)*d} \}$. Here, a total of $f$ frames are sampled and $d=\floor*{\frac{T}{f}}$. We conduct the sampling for three reasons: 1) the original video $\bm{V}$ has many frames (average $T=71$ for VSPW~\cite{miao2021vspw}), and it is unaffordable to explicitly extract global temporal contexts from all $T$ frames; 2) the contents in nearby frames have already been modeled by CFFM through static and motional contexts; 3) most contents in nearby frames are redundant and the sampled $\bar{\bm{V}}$ contains enough contexts from a global temporal view. The corresponding extracted features for the sampled frames are $\{\bm{F}_{1}, \bm{F}_{1+d}, \cdots, \bm{F}_{1+(f-1)*d} \}$.

Next, we tokenize the extracted features and treat the feature vector at each pixel as a token, resulting in tokens $\bm{o} \in \mathbb{R}^{N_o \times c}$, where $N_o=fhw$ is the total number of tokens. Those tokens contain the global temporal contexts for the whole video. However, it is impractical to exploit $\bm{o}$ due to the large number of tokens. Inspired by prototype learning~\cite{yang2021mining, li2021adaptive}, we exploit unsupervised clustering to extract typical contextual prototypes from $\bm{o}$. This largely reduces the number of tokens for the following processing, thus saving computational resources. The extracted prototypes still contain the necessary and relevant contexts for the whole video, while having a much smaller size and more condensed information. In our experiments, we use $k$-means to generate contextual prototypes $\bm{p} \in \mathbb{R}^{N_p \times c}$ from $\bm{o} \in \mathbb{R}^{N_o \times c}$, where $N_p \ll N_o$. In our experiments, $N_p$ is set to 100 unless otherwise specified.

Due to the selection of video frames across the whole video and the use of GPU-based $k$-means clustering, the process of generating global temporal contextual prototypes is fast and does not significantly decrease the speed, which will be shown in our experiments (\secref{exp:comparison}).

\subsection{Global Temporal Context Mining}
After obtaining global temporal contextual prototypes $\bm{p}$, we again exploit the CFM module to mine the global temporal contexts for refining the features of the target frame. $\bm{F}_{t}'$ is input to the first layer of the CFM module and we re-write it as $\bm{G}^{0}_{t}=\bm{F}_{t}'$. Specifically, for the $i$-th window partition in $\bm{G}^{0}_{t}$ ($\bm{F}_{t}'$), the query $Q_i$, key $K_i$, and value $V_i$ are computed using three fully connected layers as follows:
\begin{align}\label{Eq:gene_qkv_gtc}
% \footnotesize
\begin{split}
    Q_i={\rm FC}(\bm{G}^{0}_{t}[i]),~~~K_i={\rm FC}(\bm{p}),~~~V_i={\rm FC}(\bm{p}).
\end{split}
\end{align}
Here, the contextual prototypes $\bm{p}$ contain the contexts from the whole video and are shared for all the patches of the target frame in the video. Next, we use non-self attention to update the target frame features, as follows:
\begin{align}\label{Eq:gene_attn_gtc}
% \footnotesize
\begin{split}
    \bm{G}^{1}_{t}[i]=\text{Softmax}(\frac{Q_i K_i^{T}}{\sqrt{c}})V_i+\bm{G}^{0}_{t}[i],
\end{split}
\end{align}
Similar to CFFM (\secref{sec:mining}), \equref{Eq:gene_qkv_gtc} and \equref{Eq:gene_attn_gtc} are repeated for $N_g$ times. After refined by the global temporal contextual prototypes, the final feature for the target frame is given by $\bm{G}^{N_g}_{t}$, which is reshaped into $\mathbb{R}^{h \times w \times c}$.

To generate segmentation predictions, a simple MLP is used to project $\bm{G}^{N_g}_{t}$ into segmentation logits $\hat{\bm{R}}_{t}$. During training, the cross entropy loss (CE) is computed between $\hat{\bm{R}}_{t}$ and $\bm{S}_t$. During inference, we combine the logits learned from local and global temporal contexts in a weighted manner, \ie, $\bar{\bm{R}}_{t}=\lambda \hat{\bm{R}}_{t}$$+$${\bm{R}_{t}}$. In our experiment, we set $\lambda$ to be 0.5. The method using predictions given by $\bar{\bm{R}}_{t}$ is denoted as CFFM++, which is the extended version of CFFM by additionally exploiting global temporal contexts.

\begin{table*}[!t]
\centering
\setlength{\tabcolsep}{3.5mm}
\caption{\textbf{Comparison with state-of-the-art methods on the VSPW~\cite{miao2021vspw} validation set.} Our models outperform the compared methods, with better balance in terms of model size, accuracy, latency, and speed. Both FPS and MACs are computed with the input size of $480 \times 853$.}
%\resizebox{\linewidth}{!}{%
\begin{tabular}{l|c|c|c|c|c|c|c|c}
\hline
Methods    & Backbone   & Params (M)  $\downarrow$     & mIoU $\uparrow$     & Weighted IoU  $\uparrow$        & mVC$_8$  $\uparrow$    & mVC$_{16}$      $\uparrow$      &     MACs (G) $\downarrow$    & FPS (f/s) $\uparrow$   \\ \hline \hline
     % \rotatebox[origin=c]{90}
 SegFormer~\cite{xie2021segformer}  & MiT-B0                &       3.8          & 32.9                     & 56.8                                   & 82.7                      & 77.3             &    14.6     & 73.4 \\ \cline{1-9} 
 SegFormer~\cite{xie2021segformer}  & MiT-B1                &  13.8 & {36.5} & 58.8                  & {84.7}     & {79.9}   & 33.0 &  58.7 
    \\ \cline{1-9} 
CFFM (Ours)      & MiT-B0       & 4.7 & {35.4} & 58.5                  & {87.7} & {82.9} & 25.6 & 43.1  \\ \cline{1-9} 
 CFFM (Ours)  & MiT-B1     &     15.5     & {38.5}    & {60.0}    & {88.6}   & {84.1}     & 55.4 & 29.8   \\ \cline{1-9} 
 CFFM++ (Ours)  & MiT-B0       & 5.7 & 35.9 & 58.9 &  88.4   & 83.8  & 34.3 & 40.4  \\ \cline{1-9} 
 CFFM++ (Ours)  & MiT-B1     &  16.5  &  \textbf{39.9}  & \textbf{60.7}  & \textbf{89.1}  &  \textbf{84.9} & 64.2 & 27.6  \\ \hline \hline
 DeepLabv3+~\cite{chen2018encoder} & ResNet-101            &      62.7          & 34.7                      & 58.8                                & 83.2                  & 78.2          &      -      &   -   \\ \cline{1-9} 
UperNet~\cite{xiao2018unified}    & ResNet-101            &    83.2     & 36.5                      & 58.6                                   & 82.6                      & 76.1           &      -     &     - \\ \cline{1-9} 
PSPNet~\cite{zhao2017pyramid}     & ResNet-101            &       70.5       & 36.5                      & 58.1                                 & 84.2                      & 79.6            &    -      &   13.9   \\ \cline{1-9} 
OCRNet~\cite{yuan2020object}     & ResNet-101            &     58.1      & 36.7                      & 59.2                                 & 84.0                      & 79.0            &    -      &   14.3   \\ \cline{1-9} 
ETC~\cite{liu2020efficient}        & PSPNet            &    89.4      & 36.6                      & 58.3                               & 84.1                 & 79.2            &      -         &   -   \\ \cline{1-9} 
NetWarp~\cite{xiao2018unified}    & PSPNet                &   89.4    & 37.0                      & 57.9                                & 84.4                   & 79.4           &       -       &   -   \\ \cline{1-9} 
ETC~\cite{liu2020efficient}        & OCRNet                &  58.1    & 37.5                      & 59.1                                & 84.1                      & 79.1   &  -   &   -   \\ \cline{1-9} 
NetWarp~\cite{xiao2018unified}    & OCRNet                &   58.1    & 37.5                      & 58.9                                 & 84.0                      & 79.0  & - &   -   \\ \cline{1-9} 
TCB$_\text{st-ppm}$~\cite{miao2021vspw}    &   ResNet-101     & 70.5   &      37.5       &     58.6        &     87.0    &  82.1    & - &  10.0   \\ \cline{1-9} 
TCB$_\text{st-ocr}$~\cite{miao2021vspw}    &  ResNet-101     &   58.1    &     37.4       &     59.3      &   86.9      &  82.0    &- &  5.5  \\ \cline{1-9} 
TCB$_\text{st-ocr-mem}$~\cite{miao2021vspw}    &  ResNet-101      &   58.1  &   37.8    &   59.5       &      87.9          &     84.0      & - & 5.5     \\ \cline{1-9} 
SegFormer~\cite{xie2021segformer}  & MiT-B2   & 24.8 &   43.9   & 63.7 &   86.0  &   81.2 & 57.2 &   39.2   \\ \cline{1-9} 
SegFormer~\cite{xie2021segformer}  & MiT-B5   & 82.1 &  48.2    & 65.1 &   87.8  &   83.7 & 187.0 &   17.2   \\ \cline{1-9} 
CFFM (Ours)   & MiT-B2  & 26.5 &  {44.9}    & {64.9} &    {89.8}  &   {85.8}   & 79.6 & 23.8   \\ \hline
CFFM (Ours)   & MiT-B5  & 85.5 &  {49.3}    & {65.8} &   \textbf{90.8}   &   {87.1}   &232.2 & 11.3    \\ \hline
CFFM++ (Ours)   & MiT-B2  & 28.5 &   45.5   & 64.7 &   90.2   & 86.4     & 96.9  & 21.5  \\ \hline
CFFM++ (Ours)   & MiT-B5  & 87.5 &  \textbf{50.1}  & \textbf{66.5} &  \textbf{90.8}   &   \textbf{87.4}  & 249.5 & {10.4}   \\ \hline
\end{tabular}%}
\label{table:results_sota}
\end{table*}

\begin{table}[!t]
\centering
\setlength{\tabcolsep}{1.5mm}
\caption{\textbf{Comparison with state-of-the-art methods on the VSPW~\cite{miao2021vspw} test set.} Our model outperforms the compared methods. $^*$ means the test results are from~\cite{miao2021vspw}.}
\resizebox{\linewidth}{!}{%
\begin{tabular}{l|c|c|c|c|c}
\hline
Methods     & Backbone    & Params (M)     & mIoU  & mVC$_8$   & mVC$_{16}$ \\ \hline \hline
                                % \rotatebox[origin=c]{90}
 SegFormer~\cite{xie2021segformer}  & MiT-B0    &  3.8    &   30.9    & 81.6   &     75.7         \\ \hline
 SegFormer~\cite{xie2021segformer}  & MiT-B1    &  13.8   & 33.5   &   82.6     &    76.9      \\ \hline
CFFM (Ours)      & MiT-B0       & 4.7 &  {31.8}   &    {86.3}    &   {80.9}  \\ \hline 
 CFFM (Ours)  & MiT-B1   &    15.5   & {35.1}    &   {87.2}     &    {82.2}    \\ \hline 
 CFFM++ (Ours)  & MiT-B0  & 5.7 & 32.8 & 87.4 & 82.4    \\ \hline
 CFFM++ (Ours)  & MiT-B1     &  16.5  & \textbf{36.0}  & \textbf{87.9} &  \textbf{83.2} \\ \hline \hline
 DeepLabv3+$^*$~\cite{chen2018encoder} & ResNet-101   &   62.7   &  32.2  &   81.0 &  75.0    \\ \hline
UperNet$^*$~\cite{xiao2018unified}    & ResNet-101   &    83.2   &  33.5   &  79.3   &  73.3    \\ \hline
PSPNet$^*$~\cite{zhao2017pyramid}     & ResNet-101  &   70.5   & 33.8  & 83.4    &   78.3   \\ \hline
OCRNet$^*$~\cite{yuan2020object}     & ResNet-101   &   58.1  &  34.0  & 82.9   &  77.4   \\ \hline
ETC$^*$~\cite{liu2020efficient}        & PSPNet    &   89.4   &  33.8  &  82.8  &     77.1       \\ \hline
NetWarp$^*$~\cite{xiao2018unified}    & PSPNet  &   89.4    &   33.7  &   82.6 &  77.1      \\ \hline
ETC$^*$~\cite{liu2020efficient}        & OCRNet   &  58.1    &  34.6  &  83.1   &   78.0    \\ \hline
NetWarp$^*$~\cite{xiao2018unified}    & OCRNet   &   58.1    &  35.0  &  83.2  &   77.2     \\ \hline
TCB$_\text{st-ppm}$$^*$~\cite{miao2021vspw}    &   ResNet-101  & 70.5   & 34.6   &  85.2  &  80.2    \\ \hline
TCB$_\text{st-ocr}$$^*$~\cite{miao2021vspw}    &  ResNet-101  &   58.1    &  35.1  &  85.1  & 80.1    \\ \hline
TCB$_\text{st-ocr-mem}$$^*$~\cite{miao2021vspw}    &  ResNet-101  &   58.1  &  35.6  &  86.2  &  81.9     \\ \hline
SegFormer~\cite{xie2021segformer}  & MiT-B2   & 24.8 & 40.0    &   84.9     &    79.8    \\ \hline
CFFM (Ours)   & MiT-B2  & 26.5 &    {41.0}   &   {88.4}     &    {83.6}   \\ \hline
% CFFM (Ours)   & MiT-B5  & 85.5 &     &  &     \\ \hline
CFFM++ (Ours)   & MiT-B2  & 28.5 &  \textbf{42.0}  & \textbf{88.9} &   \textbf{84.7}   \\ \hline
\end{tabular}}
\label{table:results_sota_test}
\end{table}

\section{Experiments}\label{sec:experi}
\subsection{Experimental Setup}
\noindent\textbf{Implementation details for CFFM.}\quad
We implement our approach based on the \texttt{mmsegmentation}~\cite{mmseg2020} codebase and conduct all experiments on 4 NVIDIA RTX 6000 GPUs (24G memory). The backbones are the same as SegFormer~\cite{xie2021segformer}, which are all pretrained on ImageNet~\cite{russakovsky2015imagenet}. For other parts of our model, we adopt random initialization. Our model uses 3 reference frames unless otherwise specified, and we have $\{k_1, k_2, k_3\}=\{9, 6, 3\}$, following \cite{miao2021vspw}. We found that this selection of reference frames is enough to model \textit{local temporal contexts} and achieve impressive performance. 
% For the receptive field, pooling kernel, and window size, we set $r=\{49,20,6,7\}$, $p=\{7,5,3,1\}$, and $s=7$. 
For the receptive field, pooling kernel, and window size, we set $r=\{49,20,6,7\}$, $p=\{7,4,2,1\}$, and $s=7$. For the target frame, we additionally have $r_t'=35$ and $p_t'=5$.
During training, we adopt augmentations including random resizing, flipping, cropping, and photometric distortion. We use the crop size of $480 \times 480$ for the VSPW dataset \cite{miao2021vspw} and $512 \times 1024$ for Cityscapes~\cite{cordts2016cityscapes}. For optimizing parameters, we use the AdamW and ``poly'' learning rate schedule with an initial learning rate of 6$e$-5. The network is trained for 160k iterations, following SegFormer~\cite{xie2021segformer}. During testing, we conduct single-scale testing and resize all images on VSPW to the size of $480\times 853$ and $512\times 1024$ for Cityscapes. Note that for efficiency and simplicity, the predicted mask is obtained by feeding the whole image to the network, rather than using the sliding window as in~\cite{SETR}. We do \textit{not} use any post-processing such as CRF~\cite{krahenbuhl2011efficient}.

\myPara{Impelmentation details for CFFM++.} Our CFFM++ is built on CFFM. Once CFFM is trained, the corresponding encoder has the ability to extract informative features from video frames. Hence, we use the trained encoder from CFFM as the feature extractor for generating the global temporal contextual prototypes (\secref{sec:gtc_prototypes}). When generating prototypes, we set the number of sampled video frames ($f$) as 10 for all videos. The number of prototypes $N_p=100$. When mining the global temporal contexts, we set $N_g$ as 1 for small models (MiT-B0 and MiT-B1) and 2 for large models. During the training of CFFM++, we freeze the encoder and CFFM parameters, while only updating the multi-head non-self attention modules since CFFM is already well-optimized. This also reduces the training iterations needed for fine-tuning the newly added non-self attention modules. For this fine-tuning, we only use 40k iterations. The learning rate is set as 2e-4. Other settings are kept the same as CFFM.

\myPara{Datasets.}
Our experiments are mainly conducted on the VSPW dataset~\cite{miao2021vspw}, which is the largest VSS benchmark. Its training, validation, and test sets have 2,806 clips (198,244 frames), 343 clips (24,502 frames), and 387 clips (28,887 frames), respectively. It contains diverse scenarios including both indoor and outdoor scenes, annotated for 124 categories. More importantly, VSPW has dense annotations with a high frame rate of 15fps, making itself the best benchmark for VSS till now. In contrast, previous datasets used for VSS only have very sparse annotation, \ie, only one frame out of many consecutive frames is annotated. Both training and validation sets of VSPW are publicly available while the test set is not open. However, the test performance can be obtained from the VSPW2021 challenge server. On the server, the test is split into the development part and the final part, and only evaluation on the final part is available. We obtain the performance on the test set through the server. In addition to VSPW, we also evaluate the proposed method on the Cityscapes dataset~\cite{cordts2016cityscapes}, which annotates one frame out of every 30 frames.

\myPara{Evaluation metrics.}
Following previous works~\cite{shelhamer2017fully}, we use mean IoU (mIoU), and weigheted IoU to evaluate the segmentation performance. In addition, we also adopt video consistency (VC)~\cite{miao2021vspw} to evaluate the smoothness of the predicted segmentation maps in the temporal domain. Formally, for a video clips $\{\bm{I}_t\}_{t=1}^{T}$ with ground-truth segmentation masks $\{\bm{S}_t\}_{t=1}^{T}$ and predicted masks $\{\bm{S}_{t}'\}_{t=1}^{T}$, VC$_n$ is computed as follows,
\begin{align}\label{Eq:comp_vc}
% \footnotesize
\begin{split}
        \text{VC}_{n}=\frac{1}{T-n+1}\sum_{i=1}^{T-n+1}\frac{(\cap_{i}^{i+n-1}\bm{S}_i)\cap (\cap_{i}^{i+n-1}\bm{S}_i')}{\cap_{i}^{i+n-1}\bm{S}_i},
\end{split}
\end{align}
where $T\geq n$. After computing VC$_n$ for every video, we obtain the mean of VC$_n$ for all videos as mVC$_n$. The purpose of this metric is to evaluate the level of consistency in the predicted masks among those common areas (pixels' semantic labels do not change) across long-range frames. For more details, please refer to \cite{miao2021vspw}. Note that, to compute the VC metric, the ground-truth masks for all frames are needed.

The details of computing FPS are as follows. The FPS is measured in mini-batches with the batch size set to 2. We keep note of the computation time $\mathcal{T}$ for processing $\mathcal{K}$ mini-batches. The FPS can be calculated by $2\mathcal{K}/\mathcal{T}$. We set the batch size to 2 because this leads to high usage ($>$95\%) of GPU, which is common in this community. We computed the FPS for all methods in the same way for fair comparisons.

\subsection{Comparison with State-of-the-art Methods} 
\label{exp:comparison}
\myPara{Results on CFFM.}
We compare the proposed method with state-of-the-art VSS methods on VSPW~\cite{miao2021vspw} in \tabref{table:results_sota}. The results are analyzed from different aspects. For small models (number of parameters less than 20M), CFFM outperforms corresponding baselines with a clear margin, while introducing limited model complexity. For example, using the backbone MiT-B0, CFFM has 2.5\% mIoU gain over the strong baseline of SegFormer~\cite{xie2021segformer}, with the cost of increasing the parameters from 13.8M to 15.5M, increasing MACs from 33.0G to 55.4G, and reducing the FPS (frames per second) from 73.4 (f/s) to 43.1 (f/s). Our method also provides much more consistent predictions for the videos, outperforming the baseline with 5.0\% and 5.6\% in terms of mVC$_8$ and mVC$_{16}$, respectively. Note that both metrics mVC$_8$ and mVC$_{16}$ provide an evaluation of visual consistency within predicted masks for videos, as verified in~\cite{miao2021vspw}.

For large models (number of parameters $>$ 20M), CFFM achieves state-of-the-art performance in this challenging dataset and also generates visually consistent results. Specifically, our model using MiT-B2 has 26.5M parameters (slightly larger than SegFormer~\cite{xie2021segformer}) and achieves 44.9\% mIoU at the FPS of 23.8 (f/s), using 79.6G MACs. Our large model (based on MiT-B5) achieves mIoU of 49.3\% and performs best in terms of visual consistency, with mVC$_8$ and mVC$_{16}$ of 90.8\% and 87.1\%, respectively. To summarize, for all backbones (MiT-B0, MiT-B1, MiT-B2, and MiT-B5), CFFM clearly outperforms the corresponding baseline, showing that the proposed modules are stable and provide consistent performance improvement. The results validate the effectiveness of the proposed coarse-to-fine feature assembling (CFFA) and cross-frame feature mining (CFM) in mining relevant information (\textit{local temporal contexts}) from nearby frames.

\begin{table}[!t]
\centering
\setlength{\tabcolsep}{2.5mm}
\caption{\textbf{Comparison with recent efficient VSS methods on the Cityscapes~\cite{cordts2016cityscapes} dataset.} Our methods are superior to the compared methods.}
\resizebox{\columnwidth}{!}{%
\begin{tabular}{l|c|c|c|c}
\hline
Methods & Backbone   & Params (M) & mIoU  & FPS (f/s)  \\ \hline \hline
FCN~\cite{shelhamer2017fully}      & MobileNetV2    &    9.8   & 61.5    & 14.2 \\ \hline
CC~\cite{shelhamer2016clockwork}      & VGG-16     &    -   & 67.7   & 16.5 \\ \hline
DFF~\cite{zhu2017deep}     & ResNet-101 &   -    & 68.7   & 9.7  \\ \hline
GRFP~\cite{nilsson2018semantic}     & ResNet-101 &     -   & 69.4   & 3.2  \\ \hline
PSPNet~\cite{zhao2017pyramid}    & MobileNetV2 &  13.7   & 70.2   &  11.2  \\ \hline
DVSN~\cite{xu2018dynamic}     & ResNet-101 &  -     & 70.3   & 19.8 \\ \hline
Accel~\cite{jain2019accel}    & ResNet-101 &  -     & 72.1   & 3.6  \\ \hline
ETC~\cite{liu2020efficient}    & ResNet-18 &  13.2     & 71.1   & 9.5  \\ \hline \hline
SegFormer~\cite{xie2021segformer}    & MiT-B0 &  3.7  &   71.9  & 58.5  \\ \hline 
CFFM (Ours)    & MiT-B0 &   4.6    & {74.0}  & 34.2 \\ \hline
CFFM++ (Ours)    & MiT-B0 &   5.1    & 74.3 & 28.8 \\ \hline \hline
SegFormer~\cite{xie2021segformer}    & MiT-B1 & 13.8   &   74.1  &  46.8 \\ \hline 
CFFM (Ours)    & MiT-B1 &    15.4   & {75.1}  & 23.6 \\ \hline
CFFM++ (Ours)    & MiT-B1 &   15.9    & \textbf{75.7}  & 20.4 \\ \hline
\end{tabular}}
\label{table:results_city}
\end{table}

We also obtain results on the test set of the VSPW dataset from the VSPW2021 challenge server, which is shown in \tabref{table:results_sota_test}. We can observe that the proposed CFFM surpasses the considered approaches. For example, upon MiT-B1, CFFM is clearly better than the baseline (SegFormer), with an mIoU gain of 1.6\%.
The experimental results on the Cityscapes~\cite{cordts2016cityscapes} dataset are shown in \tabref{table:results_city}. Our method is compared with recent efficient segmentation methods. Only using 4.6M parameters, CFFM obtains 74.0\% mIoU with an FPS of 34.2 (f/s), achieving an excellent balance on model size, accuracy, and speed. When using a deeper backbone, we achieve 75.1\% mIoU with an FPS of 23.6 (f/s). Note that this dataset has sparse annotations, the excellent performance demonstrates that our method works well for both fully supervised and semi-supervised settings.

\myPara{Results on CFFM++.} CFFM++, the extension of CFFM by additionally exploring \textit{global temporal contexts}, achieves consistent improvements over CFFM on the VSPW dataset under all studied backbones while introducing limited computation resources. For example, using MiT-B1, CFFM++ obtains an mIoU gain of 1.4\% over CFFM and generates more visually consistent predictions with a gain of 0.8\% in the metric mVC$_{16}$.
We also compute the average improvement of CFFM++ over CFFM across various backbones. On average, CFFM++ noticeably outperforms CFFM by 0.9 and 1.0 point on the VSPW val and test sets, respectively. The improvements are valid for all datasets and backbones, showing the effectiveness of mining global temporal contexts.

We also observe that CFFM++ is efficient in terms of model size and computation speed. For example, with MiT-B1, CFFM++ only increases the number of parameters from 15.5M to 16.5M and reduces the FPS from 29.8 (f/s) to 27.6 (f/s), compared with CFFM. The reasons why the proposed CFFM++ does not noticeably increase the latency are in three aspects. First, the number of global contextual prototypes is very small, \ie, 100 in our main experiments. Second, in CFM, only 2 non-self attention layers are used. Third, the decoder (\figref{fig:framework} and \figref{fig:framework_exten}) is very small, comprising a single convolutional layer mapping to class logits, following SegFormer~\cite{xie2021segformer}. It means that the additional latency caused by mining global temporal contexts (CFFM++) is very insignificant, compared to the computation of CFFM. Therefore, CFFM++ is only a little slower than CFFM.

\begin{figure*}[!t]
    \centering
    \includegraphics[width=1.0\linewidth]{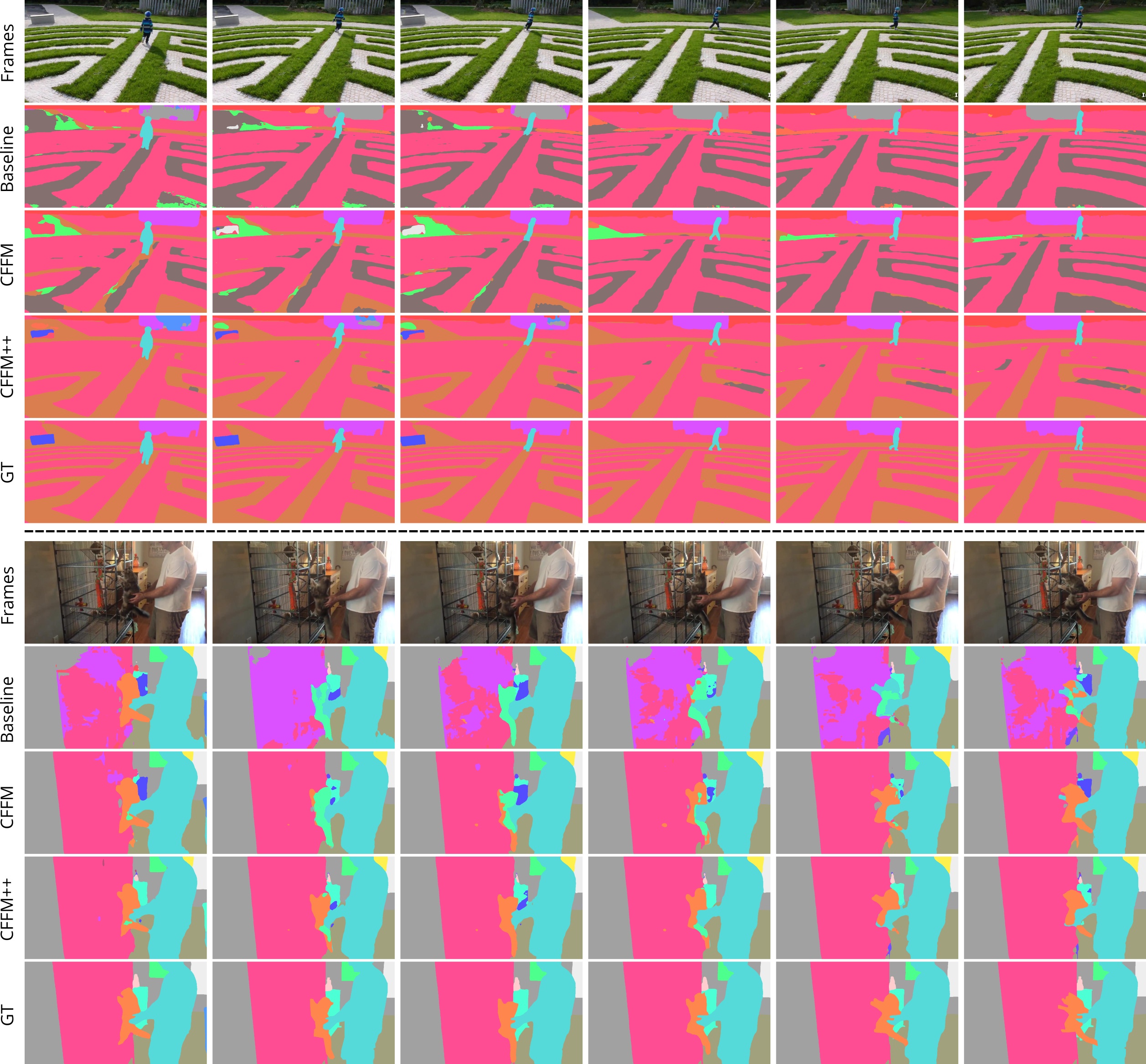}
    \caption{\textbf{Qualitative results for two video clips.} We compare the proposed methods with the baseline (SegFormer~\cite{xie2021segformer}) visually. From \textit{top} to \textit{down}: the input video frames, the predictions of SegFormer~\cite{xie2021segformer}, CFFM predictions, CFFM++ predictions and the ground truth (GT). It shows that CFFM produces more accurate and consistent results, compared to the strong baseline. Furthermore, by using global temporal contexts, CFFM++ further improves over CFFM. \textit{Best viewed in color.}}
    \label{fig:qual}
\end{figure*}

\myPara{Qualitative Results.} The qualitative results are shown in Fig.~\ref{fig:qual}. For the given examples, CFFM resolves the inconsistency existing in the predictions of the baseline, as it exploits the \textit{local temporal contexts} within nearby frames. Build upon CFFM, CFFM++ further improves the per-frame prediction accuracy and temporal consistency by utilizing the \textit{global temporal contexts} within a long range of video frames.

\begin{table}[!t]
\centering
\setlength{\tabcolsep}{2.6mm}
\caption{\textbf{Ablation study on the number of attention layers in CFM.}}
\resizebox{1.0\columnwidth}{!}{%
\begin{tabular}{l|c|c|c|c|c}
\hline
Methods                & $N$ & mIoU  & mVC$_8$ & mVC$_{16}$ & Params (M) \\ \hline \hline
\multicolumn{6}{c}{MiT-B0} \\ \cline{1-6} 
SegFormer~\cite{xie2021segformer}   & - &   32.9     & 82.7 &  77.3   &  3.8    \\ \cline{1-6} 
\multirow{2}{*}{CFFM (Ours)}    & 1 &       35.4         &    \textbf{87.7}  &  82.9   &  4.7        \\ \cline{2-6} 
   & 2 &      \textbf{35.7}        &  \textbf{87.7}  &  \textbf{83.0}   &  5.5     \\ \hline \hline
\multicolumn{6}{c}{MiT-B1} \\ \cline{1-6} 
SegFormer~\cite{xie2021segformer}    & - &   36.5    & 84.7 & 79.9    &     13.8     \\ \cline{1-6} 
\multirow{4}{*}{CFFM (Ours)}    & 1 &  37.8  &  88.3    & 83.6     & 14.6       \\ \cline{2-6} 
   & 2 &    38.5    &  \textbf{88.6}  &  \textbf{84.1}  &     15.5    \\ \cline{2-6} 
   & 3 &       38.7    &  \textbf{88.6}   &  \textbf{84.1}   &   16.3    \\ \cline{2-6} 
   & 4 &       \textbf{38.8}        &   88.5   & 83.9      &   17.2         \\ \hline
\end{tabular}}
\label{table:albation_n}
\end{table}

\begin{table}[!t]
\centering
\setlength{\tabcolsep}{3.mm}
\caption{\textbf{Ablation study on the selection of the reference frames.} We use MiT-B1 as the backbone.}
\resizebox{1.0\columnwidth}{!}{%
\begin{tabular}{l|c|c|c|c|c|c}
\hline
Methods   & k1 & k2 & k3 &  mIoU & mVC$_8$ & mVC$_{16}$ \\ \hline
SegFormer             & -  & -  & -  &   36.5    &   84.7 &   79.9  \\ \hline \hline
\multirow{5}{*}{CFFM (Ours)} & -  & -  & 3  &  37.4 &  87.4   &   82.4 \\ \cline{2-3} \cline{4-7} 
    & -  & -  & 6  &   37.7   &   88.0   &    83.3   \\ \cline{2-3} \cline{4-7} 
    &   -  & -  & 9  &  37.9    &  88.4   &       83.9 \\ \cline{2-3} \cline{4-7} 
     & 3  & 2  & 1  &    37.7    & 88.3  &  83.6           \\ \cline{2-3} \cline{4-7} 
    & 9  & 6  & 3  &     \textbf{38.5}  & \textbf{88.6} &  \textbf{84.1}          \\ \hline
\end{tabular}}
\label{table:albation_reference}
\end{table}

\subsection{Ablation Study}
All ablation studies are conducted on the large-scale VSPW~\cite{miao2021vspw} dataset and follow the same training strategies as described above, for a fair comparison.

\myPara{Influence of the number of attention layers.}
\tabref{table:albation_n} shows the performance of CFFM with respect to the number of non-self attention layers in the CFM module. For two backbones of MiT-B0~\cite{xie2021segformer} and MiT-B1~\cite{xie2021segformer}, CFFM clearly outperforms the corresponding baseline (SegFormer) when using only a single attention layer and introducing a small number of additional parameters. It demonstrates the effectiveness of the proposed CFFA module and the non-self attention layer. The former efficiently extracts the \textit{local temporal contexts} from the nearby frames and the latter effectively mines the contextual information to refine target frame features. In addition, we observe there is a trade-off between performance and the model complexity (number of parameters) on the MiT-B1 backbone. When using more attention layers within the CFM module, better mIoU is obtained while the model size linearly increases. For our method (CFFM) on MiT-B1, we choose $N=2$ since a better trade-off is observed.

\myPara{Impact of selection of reference frames.}
We study the impact of the selection of reference frames for learning \textit{local temporal contexts} in \tabref{table:albation_reference}. We start by using a single reference frame. There seems to be a trend that when increasing the distance between the reference frame and the target frame, better performance is obtained. The reason for this is that the more faraway reference frame may contain richer and more different contexts which complements the contexts of the target frame. This also suggests that extracting global temporal contexts from the whole video is useful. When using more reference frames ($k_1=9, k_2=6, k_3=3$), the best performance with mIoU of 38.5\% is achieved. It is worth noting that CFFM using reference frames combination of $k_1=3, k_2=2$, and $k_3=1$, only achieves segmentation mIoU of 37.7\% and performs similarly as the cases when using a single reference frame. It is possibly due to the fact that the very close reference frames do not give much new information for segmenting the target frame, as also shown in~\cite{miao2021vspw}.

\myPara{Impact of CFFA and CFM.} Starting from SegFormer~\cite{xie2021segformer}, we only add CFFA to extract contextual tokens. To mine the generated contexts, we use an MLP to process them, which are finally merged with the target features. For this variant of only using CFFA, we obtain a mIoU of 37.6\%, outperforming the baseline with a mIoU of 36.5\% by 1.1\% gain. Then, we add both CFFA and CFM on top of the baseline, which is our final model (CFFM) for learning \textit{local temporal contexts}. The segmentation performance (mIoU) for CFFM is 38.5\%. These facts verify that both CFFA and CFM modules are valuable and essential to the proposed CFFM mechanism for learning local temporal contexts.

\myPara{Impact on the Receptive Fields.}
To investigate the impact of receptive fields, we conduct ablation study on different $r=\{r_{t-k_1},r_{t-k_2},r_{t-k_3},r\}$. As mentioned previously, we use three reference frames, with $k_1=9$, $k_2=6$, and $k_3=3$, following \cite{miao2021vspw}. Note that larger receptive fields mean that the larger region/context is used by the network and more context tokens are generated, leading to more computational cost in the proposed CFM module. For fairness, when studying the impact of $r$, we keep other parameters unchanged. The ablation study is conducted on the VSPW~\cite{miao2021vspw} validation set.

The results are shown in Tab.~\ref{table:ablation_r}. We have several observations. First, when using small receptive fields, our method achieves the mIoU of 37.2\%, which is already better than the baseline (SegFormer~\cite{xie2021segformer}) with the mIoU of 36.5\%. Second, when increasing the receptive fields so that the model can see larger regions in farther frames, the performance significantly improves, from 37.2\% to 39.2\%, implying the value of static and motional contexts. Third, further increasing the receptive fields to $\{49,35,21,7\}$ does not boost performance. The possible reason is that when receptive fields become large enough, no further useful contexts can be exploited. In general, using a reasonable $r$ gives good performance and our method is robust to the reasonable choice of the receptive fields $r$.

\myPara{Impact on the Pooling Windows.}
To investigate the impact of pooling kernels/windows, we conduct ablation study on different $p=\{p_{t-k_1},p_{t-k_2},p_{t-k_3},p_t\}$, where $k_1=9$, $k_2=6$, and $k_3=3$, following \cite{miao2021vspw}. While ablating $p$, we keep other parameters the same for fair comparisons. Note that a smaller pooling window indicates more fine-grained features are extracted, and hence more context tokens are generated, leading to more computational cost in our multi-head non-self attention layer. The ablation study is conducted on the VSPW~\cite{miao2021vspw} validation set.

The results are shown in Tab.~\ref{table:ablation_p}. First, for different choices of pooling windows $p$, our method is much better than the SegFormer~\cite{xie2021segformer} baseline with mIoU of 36.5\%.  When increasing the granularity (more fine-grained features are exploited) from $\{7,7,7,1\}$ to $\{7,5,3,1\}$, and to $\{5,3,3,1\}$, better mIoU scores are obtained, \ie, from 38.3\% to 38.5\%, and to 38.7\%. In general, our method is robust to the choice of $p$ and the reasonable $p$ gives a good performance.

\begin{table}[!t]
\centering
\caption{\textbf{Ablation study on the impact of the receptive fields.} The used backbone is MiT-B1. The proposed method is robust to reasonable receptive fields.}
\setlength{\tabcolsep}{6.0mm}
\resizebox{\columnwidth}{!}{%
\begin{tabular}{c|c|c|c}
\hline
$r$              & mIoU & mVC$_8$ & mVC$_{16}$ \\ \hline
\{7,5,3,7\}    &   37.2   & 88.0   &  83.4     \\ \hline
\{21,15,9,7\}  &   38.0   &   88.2  &    83.6     \\ \hline
\{35,15,9,7\}  &   38.3   &   88.2  &    83.7     \\ \hline
\{49,15,9,7\}  &  38.7    &   88.0     &    83.3     \\ \hline
\{49,20,6,7\}  &    38.5  &     88.6   &    84.1     \\ \hline
\{49,25,15,7\} &   \textbf{39.2}   &    \textbf{88.6}    &    \textbf{84.1}     \\ \hline
\{49,35,21,7\} &   38.8   &   88.6     &    84.1     \\ \hline
\end{tabular}}
\label{table:ablation_r}
\end{table}

\begin{table}[!t]
\centering
\caption{\textbf{Ablation study on the impact of the pooling windows.} The used backbone is MiT-B1. The proposed method is robust to the choice of pooling windows.}
\setlength{\tabcolsep}{6.5mm}
\resizebox{\columnwidth}{!}{%
\begin{tabular}{c|c|c|c}
\hline
$p$         & mIoU & mVC$_8$ & mVC$_{16}$ \\ \hline
\{5,3,3,1\} &   \textbf{38.7}   &   88.2     & 83.7        \\ \hline
\{7,5,3,1\} &   38.5   &    \textbf{88.6}    &   \textbf{84.1}      \\ \hline
\{7,7,7,1\} &   38.3   &    88.3    &    83.8    \\ \hline
\end{tabular}}
\label{table:ablation_p}
\end{table}

\begin{table*}[!t]
\centering
\setlength{\tabcolsep}{6.6mm}
\caption{\textbf{Ablation study on local temporal contexts (static and motional contexts) and global temporal contexts.}}
\begin{tabular}{c|l|c|c|c}
\hline
Symbol & Methods       & mIoU & mVC$_8$   & mVC$_{16}$ \\ \hline
A & Baseline                          &  36.5    &  84.7     &   79.9     \\ \hline
B & Baseline+static contexts          &   37.7   &  84.4     &  79.4      \\ \hline
C & Baseline+static/motional contexts (CFFM) &  38.5    &  88.6     &  84.1      \\ \hline
D & Baseline+global temporal contexts          &   38.4   &   85.0    &   80.3    \\ \hline
E & Baseline+static contexts+global temporal contexts           &  39.2    &  85.5     & 80.8       \\ \hline
F & \begin{tabular}[c]{@{}l@{}}Baseline+static/motional contexts\\+global temporal contexts (CFFM++)\end{tabular}        &   \textbf{39.9}   &   \textbf{89.1}    &  \textbf{84.9}      \\ \hline
\end{tabular}
\label{table:albation_static_motional}
\end{table*}

\begin{table}[!t]
\centering
\setlength{\tabcolsep}{3.5mm}
\caption{\textbf{Ablation study on the number of input frames.}}
\resizebox{\columnwidth}{!}{%
\begin{tabular}{c|l|c|c|c}
\hline
$N_f$                 &  Methods  &   ~~mIoU~~ & ~~mVC$_8$~~   & ~~mVC$_{16}$~~ \\ \hline
\multirow{3}{*}{4}  & CFFM   &   38.5   &  88.6    &  84.1     \\ \cline{2-5} 
                    & CFFM+  &   38.0   &  84.9    & 80.1      \\ \cline{2-5} 
                    & CFFM++ &   \textbf{39.6}   &  \textbf{88.8}    & \textbf{84.6}      \\ \hline
\multirow{3}{*}{6} & CFFM   &   38.8   &   89.1   &  84.4     \\ \cline{2-5} 
                    & CFFM+  &   38.2   &  85.1    & 80.2      \\ \cline{2-5} 
                    & CFFM++ &   \textbf{40.0}   &  \textbf{89.4}    &   \textbf{85.3}    \\ \hline
\end{tabular}
}
\label{table:albation_num_frames}
\end{table}

\begin{table}[!t]
\centering
\setlength{\tabcolsep}{3.5mm}
\caption{\textbf{Ablation study on the number ($N_p$) of extracted prototypes.} We use MiT-B1 as the backbone.}
\resizebox{\columnwidth}{!}{%
\begin{tabular}{l|c|c|c|c}
\hline
Methods & ~$N_p$~ & ~mIoU~ & ~mVC$_8$~ & ~mVC$_{16}$~ \\ \hline
CFFM (Ours)   & -    &   38.5  &    88.6    &  84.1       \\ \hline
\multirow{4}{*}{CFFM++ (Ours)}  & 10   &   39.5   &     88.7   &    84.5     \\ \cline{2-5}
  & 50   &   39.8   &   89.0     &  84.8       \\ \cline{2-5}
  & 100  &   \textbf{39.9}   &   \textbf{89.1}     &  \textbf{84.9}       \\ \cline{2-5}
  & 200  &   39.7   &  89.0      &  84.8       \\ \hline
\end{tabular}}
\label{table:albation_num_prototypes}
\end{table}

\begin{table}[!t]
\centering
\setlength{\tabcolsep}{3.2mm}
\caption{\textbf{Study on effectiveness of prototypes.} We use prototypes extracted from different backbone models for CFFM++ (MiT-B0).}
\resizebox{1.0\columnwidth}{!}{%
\begin{tabular}{c|c|c|c|c}
\hline
Methods & Prototypes Model & ~mIoU~ & ~mVC$_8$~ & ~mVC$_{16}$~ \\ \hline
\multirow{4}{*}{\shortstack{CFFM++ \\ (MiT-B0)}}  & MiT-B0   &  35.9   &    88.4    &  83.8       \\ \cline{2-5}
  & MiT-B1   &   38.0   &  88.7    & 84.6      \\ \cline{2-5}
  & MiT-B2  &  39.1   & 89.1     &  85.1     \\ \cline{2-5}
  & MiT-B5  &   \textbf{39.6}  &   \textbf{89.3}    & \textbf{85.4}       \\ \hline
\end{tabular}}
\label{table:albation_eff_prototypes}
\end{table}

\myPara{Ablation on local and global temporal contexts.} In this experiment, we study the impact of local temporal contexts (static and motional contexts) and global temporal contexts on performance. Different from previous methods, the proposed CFFM can learn both static and motional contexts (local temporal contexts) in a unified model. When CFFM predicts the segmentation mask for the current frame, it uses three previous frames as reference frames. Based on CFFM, we further propose CFFM++ which extends CFFM by further adding global temporal contexts. The results for this ablation study are shown in Tab.~\ref{table:albation_static_motional}. We start from ``Baseline'' method (A) which means SegFormer with MiT-B1 backbone. We simulate a case where only static contexts can be used, by replicating the current frame three times and using them as the reference frames. In this way, only static contexts could be used since all the reference frames are the same as the current one. We denote this experiment as ``Baseline+static contexts" (B). We also conduct experiments on only adding global temporal contexts on ``Baseline'', which leads to D. By adding global temporal contexts on CFFM (C), we obtain our full model CFFM++ (F). Following our setting in ablation studies, we use the VSPW val dataset.

From the table, it can be seen that by adding static contexts to the baseline (A), a mIoU gain of 2.2 is achieved. However, this
variant (B) does not show improvements in temporal consistency metrics mVC8 and mVC16 since no temporal information from neighboring frames is used. When adding static/motional contexts to the baseline, the method is CFFM (C) and achieves performance gains in all metrics mIoU, mVC$_{8}$ and mVC$_{16}$. By further adding global temporal contexts on top of CFFM, we get our full model CFFM++ (F) which outperforms CFFM, showing the effectiveness of global temporal contexts. By comparing D and A, we can also see the power of global temporal contexts.

\myPara{Ablation on the number of frames.} 
For CFFM, the number of frames being used to extract local temporal contexts is $N_f=l+1$, where $l$ is the number of reference frames as introduced in \secref{sec:ltc}. To fairly study the impact of the number of frames, we also set the hyperparameter $f$ (\secref{sec:gtc}) for extracting global temporal contexts to be $l$. Hence, the number of frames being used for local and global temporal contexts is the same. In this part, we ablate on $N_f$. We also show the result of a variant \textit{CFFM+}, which \textit{only} uses global temporal contexts. The results are shown in Tab.~\ref{table:albation_num_frames}.

From the table, it can be seen that when using the same number of frames for extracting local and global temporal contexts, CFFM++ always outperforms CFFM. This is due to the fact that CFFM++ (using local and global temporal contexts) is built on top of CFFM (using local temporal contexts). Comparing CFFM+ and CFFM, we can observe CFFM is slightly better than CFFM+, since local temporal contexts are more informative than the global temporal contexts. The fact that CFFM++ outperforms both CFFM and CFFM+ shows that local temporal contexts and global temporal contexts are complementary.

\myPara{Influence of the number of global temporal contextual prototypes.} In our experiments, we set the number ($N_p$) of contextual prototypes as 100 when extracting global temporal information. Here, we study the influence of this parameter. The results are shown in \tabref{table:albation_num_prototypes}. we observe that compressing the global temporal contexts into only 10 prototypes already gives promising results. It demonstrates the effectiveness of the global temporal contexts and that the information from faraway video frames can provide additional guidance to help segment the target frame. When increasing the number of generated prototypes from 10 to 100, better performance can be achieved due to the fact that more detailed contexts are extracted and exploited. However, further extracting more (\eg, 200) contextual prototypes does not help. For a certain video, a good number (\eg, 100) of prototypes could already represent the existing contextual information well. Further increasing the number of prototypes will not significantly include more information. This is consistent with the discovery in the few-shot semantic segmentation paper ASGNet~\cite{li2021adaptive}.

\myPara{Knowledge distillation of global temporal contextual prototypes.} Here, we conduct knowledge distillation experiments using global temporal contextual prototypes. Specifically, the prototypes (with size of $\mathbb{R}^{100 \times 256}$) extracted from large models (MiT-B5, MiT-B2, MiT-b1) are used by CFFM++ on small backbone (MiT-B0). This study can be interpreted from the perspective of knowledge distillation. We distill knowledge from large models to the small model, through the extracted contextual prototypes. The results are shown in \tabref{table:albation_eff_prototypes}. It can be observed that by simply replacing the prototypes from MiT-B0 with the prototypes from MiT-B1, MiT-B2, and MiT-B5, significant performance improvements are obtained for CFFM++ (MiT-B0), which demonstrates the extracted prototypes contain rich contextual information.

\section{Conclusion and Future Work}
The video contexts contain \textit{local temporal contexts} which represent the contextual information from neighbouring/nearby frames and \textit{global temporal contexts} which indicate the contexts from the whole video. This paper first studies {local temporal contexts} which can be further divided into \textit{static contexts} and \textit{motional contexts} within the nearby frames. Previous methods pay much attention to {motional contexts} but ignore the {static contexts}. We propose a Coarse-to-Fine Feature Mining (CFFM) technique to jointly learn a unified presentation of static and motional contexts, for precise and efficient VSS. CFFM contains two parts: Coarse-to-Fine Feature Assembling (CFFA) and Cross-frame Feature Mining (CFM). The former summarizes contextual information with different granularity for different frames, according to their distance to the target frame. The latter efficiently mines the contexts from neighbouring frames to enhance the feature of the target frame. To make use of \textit{global temporal contexts}, we further propose CFFM++ which abstracts global temporal contextual prototypes from the video by unsupervised clustering and then exploits them to improve the target frame features. Extensive experiments show that CFFM boosts segmentation performance in a clear margin while adding limited computational cost. What's more, CFFM++ clearly surpasses CFFM with the help of global temporal information.

For future work, in addition to the above two aspects (local and global temporal contexts), the following two directions are promising. First, our exploration of contextual information for VSS focuses on simultaneously learning temporal contexts for all semantic categories. Considering the relationships amongst various categories (\eg, \textit{horses} is often related to the \textit{grassland}), the explicit modeling of class-specific temporal contexts is also an interesting direction to explore. Second, it would be also interesting to extend our methods to other video tasks that require the learning of temporal contexts.

{\small
\bibliographystyle{IEEEtran}
\bibliography{reference}
}

\newcommand{\AddPhoto}[1]{\includegraphics[width=1in,keepaspectratio]{Authors/#1}}

\begin{IEEEbiography}[\AddPhoto{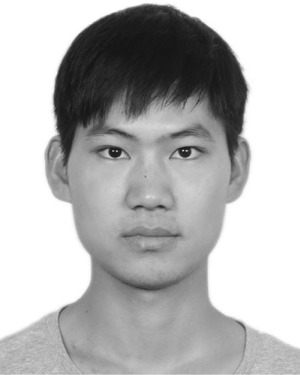}]{Guolei Sun}
received his Ph.D. degree at ETH Zurich, Switzerland, in Prof. Luc Van Gool's Computer Vision Lab in Jan 2024.  Before that, he got master degree in computer science from the King Abdullah University of Science and Technology (KAUST), in 2018. He is currently a postdoctoral researcher at Computer Vision Lab, ETH Zurich. From 2018 to 2019, he worked as a research engineer with the Inception Institute of Artificial Intelligence, UAE. His research interests include deep learning for video understanding, semantic/instance segmentation, object counting, and weakly supervised learning.
\end{IEEEbiography}
\vspace{-.2in}

\begin{IEEEbiography}[\AddPhoto{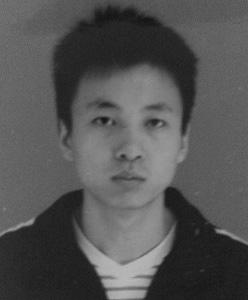}]{Yun Liu}
received his B.E. and Ph.D. degrees from Nankai University in 2016 and 2020, respectively.
Then, he worked with Prof. Luc Van Gool as a postdoctoral scholar at Computer Vision Lab, ETH Zurich, Switzerland.
Currently, he is a senior scientist at the Institute for Infocomm Research (I2R), A*STAR, Singapore.
His research interests include computer vision and machine learning.
\end{IEEEbiography}
\vspace{-.2in}

\begin{IEEEbiography}[\AddPhoto{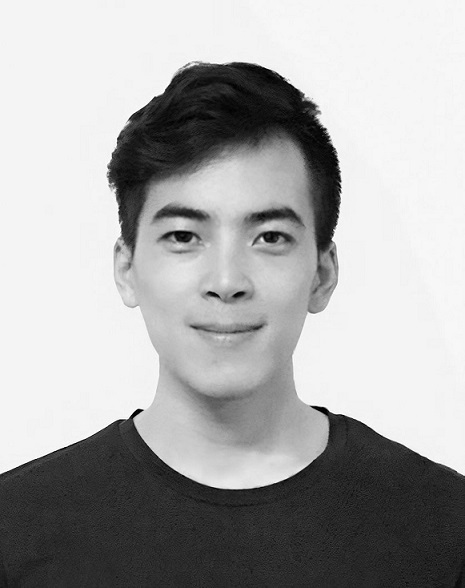}]{Henghui Ding} received his B.E. degree from Xi'an Jiaotong University, Xi'an, China, in 2016. He received the Ph.D. degree from
Nanyang Technological University (NTU), Singapore, in 2020. He was a Postdoctoral Researcher at the Computer Vision Lab of ETH Zurich in Switzerland and a Research Scientist at ByteDance AI Lab in Singapore. He is currently a Presidential Postdoctoral Fellow (Principal Investigator) at NTU. He serves as Associate Editors for IET Computer Vision and Visual Intelligence. He serves/served as Area Chair for CVPR'24 and ACM MM'24, and Senior Program Committee member for AAAI'(22-24) and IJCAI'(23-24).
His research interests include computer vision and machine learning.
\end{IEEEbiography}
\vspace{-.2in}

\begin{IEEEbiography}[\AddPhoto{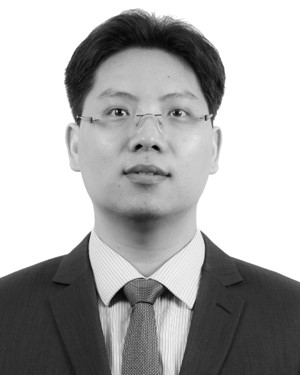}]{Min Wu}
(Senior Member, IEEE) received the B.E. degree in computer science 
from USTC, China, in 2006, 
and the Ph.D. degree in computer science from NTU, Singapore, in 2011.
He is currently a principal scientist with the Institute for Infocomm Research (I2R), A*STAR, Singapore.
He received the best paper awards in the IEEE ICIEA 2022, 
the IEEE SmartCity 2022, the InCoB 2016, and the DASFAA 2015. 
He also won the CVPR UG2+ challenge in 2021 and the IJCAI competition 
on repeated buyers prediction in 2015.
His current research interests include machine learning, data mining, 
and bioinformatics.
\end{IEEEbiography}
\vspace{-.2in}

\begin{IEEEbiography}[\AddPhoto{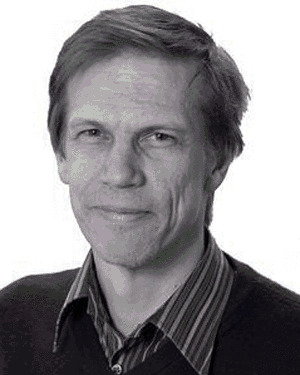}]{Luc Van Gool}
received the degree in electromechanical engineering from the Katholieke Universiteit Leuven, in 1981. Currently, he is a professor at the Katholieke Universiteit Leuven in Belgium and the ETH in Zurich, Switzerland. He leads computer vision research with both places and also teaches at both. He has been a program committee member of several major computer vision conferences. His main research interests include 3D reconstruction and modeling, object recognition, tracking, gesture analysis, and a combination of those. He received several Best Paper awards, won a David Marr Prize and a Koenderink Award, and was nominated Distinguished Researcher by the IEEE Computer Science committee. He is a co-founder of 12 spin-off companies.
\end{IEEEbiography}

\vfill

\end{document}